\documentclass[11pt,a4paper]{article}
\usepackage[hyperref]{eacl2021}
\usepackage{times}
\usepackage{latexsym}

\usepackage{microtype}

\usepackage{xcolor}
\usepackage{booktabs}
\usepackage{multirow}
\usepackage{bm}
\usepackage{arydshln}
\usepackage{float}
\usepackage{csquotes}
\usepackage{makecell}
\usepackage{caption}
\usepackage{subcaption}
\usepackage[inline]{enumitem}
\usepackage[export]{adjustbox}
\usepackage{stackengine}
\usepackage{amsmath}
\usepackage{amssymb}
\usepackage{pifont}
\usepackage{dblfloatfix}
\usepackage{verbatim}
\usepackage{pgfplots}
\usepackage{tikz}
\usetikzlibrary{arrows.meta}
\usetikzlibrary{positioning}
\usetikzlibrary{fit,backgrounds}
\usetikzlibrary{calc}

\newcommand{\vect}[1]{\mathbf{#1}}
\newcommand{\ra}[1]{\renewcommand{\arraystretch}{#1}}
\newcommand{\bftab}{\fontseries{b}\selectfont}

\newcommand{\smallcite}[1]{}

\aclfinalcopy %

\title{Large-Scale Zero-Shot Image Classification from Rich and Diverse Textual Descriptions}

\author{Sebastian Bujwid \\
  KTH Royal Institute of Technology \\
  Stockholm, Sweden \\
  \texttt{bujwid@kth.se} \\\And
  Josephine Sullivan \\
  KTH Royal Institute of Technology \\
  Stockholm, Sweden \\
  \texttt{sullivan@kth.se} \\}

\date{}

\def\NewDataset{\textit{ImageNet-Wiki}}

\begin{document}
\maketitle
\begin{abstract}
    We study the impact of using rich and diverse textual descriptions of classes for zero-shot learning (ZSL) on ImageNet. We create a new dataset \NewDataset{} that matches each ImageNet class to its corresponding Wikipedia article. We show that merely employing these Wikipedia articles as class descriptions yields much higher ZSL performance than prior works. Even a simple model using this type of auxiliary data outperforms state-of-the-art models that rely on standard features of word embedding encodings of class names.
    These results highlight the usefulness and importance of textual descriptions for ZSL, as well as the relative importance of auxiliary data type compared to algorithmic progress.
    Our experimental results also show that standard zero-shot learning approaches generalize poorly across categories of classes.
\end{abstract}

\section{Introduction}

Zero-shot learning (ZSL) relates information from different modalities (e.g., text or attributes with images), and the hope is that a sparsity of information or training data in one modality can be compensated by the other. This is important when the cost of creation or collection of training data greatly differs between the different modalities. Natural language descriptions and existing text repositories can provide this rich and practically accessible information about visual concepts and classes when no accompanying labeled images are available.

\begin{figure*}[h]
    \centering
    \includegraphics[width=1.0\linewidth]{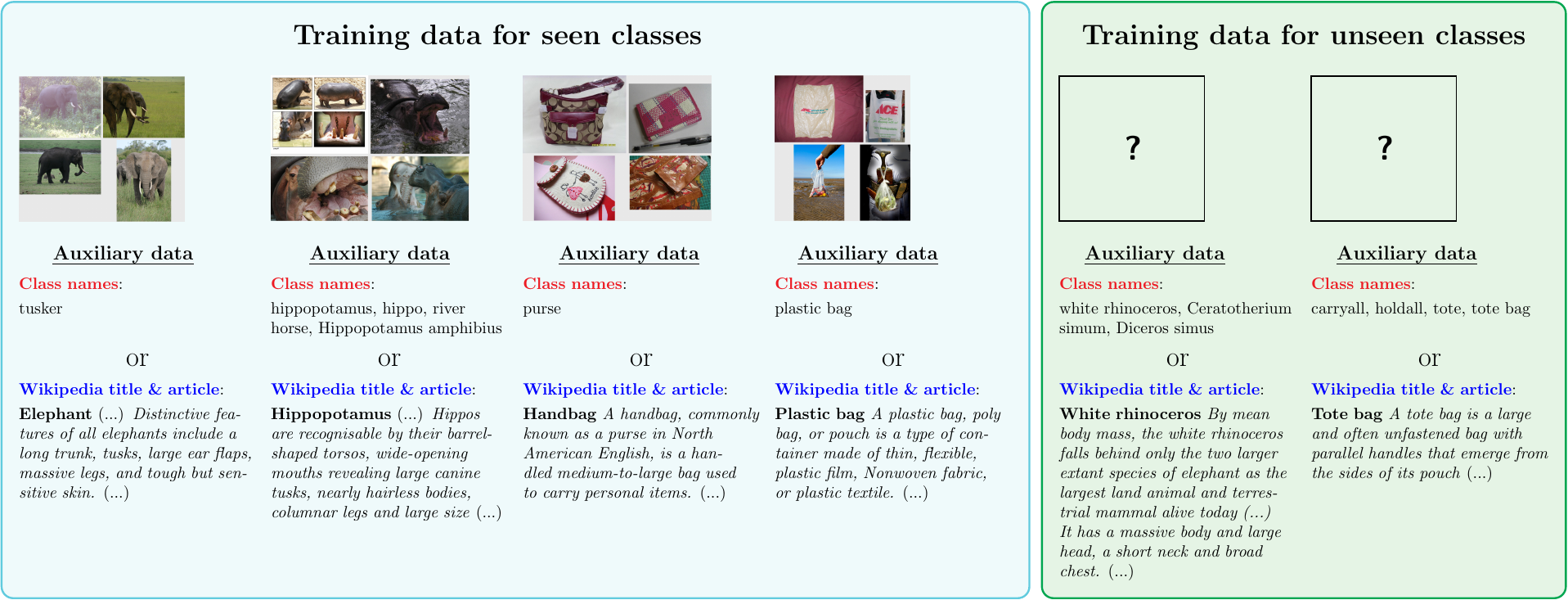}
    \caption{
       \textbf{ImageNet classes and their corresponding auxiliary data: Wikipedia articles or class names.}
       Above illustrates the difference between class names, the standard source of auxiliary data used on ImageNet, and the textual descriptions we collected.
       The short, selected article fragments shown contain information that is visually discriminative for the classes.
       In ZSL, for seen classes, we have access to both auxiliary data and images.
       However, for unseen classes only auxiliary data is available for training, with no images.
    }
    \label{fig:zsl_wiki}
\end{figure*}

Recent works on zero-shot image classification \cite{xian2019f, schonfeld2019generalized, xian2018zero} show the field's substantial progress on many of the standard ZSL benchmarks. However, most of those benchmarks mostly cover either a very small or narrow set of classes (e.g., only bird species), where human-made class attributes are often used as auxiliary data.
Unfortunately, on ImageNet, where such attributes are not available, the performance is still very low.
Therefore, in this work, rather than focus on algorithmic development, we instead study how the auxiliary data type impacts performance.
We evaluate the benefits of using textual descriptions of the ImageNet classes. We collect text from Wikipedia articles describing each class in ImageNet, illustrated in Figure~\ref{fig:zsl_wiki}.
Throughout the paper, we refer to the dataset of ImageNet classes and their corresponding Wikipedia articles, as well as the extracted text, as \NewDataset{}.\footnote{
The dataset, code and trained models available at: \url{https://bujwid.eu/p/zsl-imagenet-wiki}
}

The use of textual descriptions for ZSL has been studied before on smaller datasets~\cite{reed2016learning,elhoseiny2016write,elhoseiny2017link,zhu2018generative}.
However, this is the first work to use such textual descriptions on a large dataset, with a broad set of classes.
The major differences between commonly used datasets are highlighted in Table~\ref{tbl:zsl_datasets}.
The large-scale setup enables us to study the main challenges of a more realistic and practical zero-shot image classification scenario
and study the generalization of models to novel groups of classes (e.g., animal species in general), not only individual classes (e.g., specific animal species).

 \begin{table*}
   \caption[]{\textbf{High-level overview of the most popular existing zero-shot learning image datasets.}
   Numbers with (*) correspond only to the training set only.
   Provided by: (\dag) - \citet{reed2016learning}; (\ddag) - \citet{elhoseiny2017link}.
   }
   \label{tbl:zsl_datasets}
   \begin{center}
 {\footnotesize
 \begin{tabular}{lrrcccr}
   \toprule
   \textbf{Dataset} & \textbf{\# images} & \textbf{\# classes} & \multicolumn{3}{c}{\textbf{Used auxiliary data}} & \textbf{Semantic classes}\\[3pt]
    & & & Other & Wiki articles & Attrib. & \\[3pt]
   \midrule
   aP/aY \smallcite{Farhadi:cvpr:09} & 15k & 32
    & & & \checkmark
    & {\footnotesize person, animals, objects}\\[3pt]%
   AwA1 \smallcite{Lampert:tpami:14} & 30k & 50
    & & & \checkmark
    & {\footnotesize animals}\\[3pt]
   AwA2 \smallcite{xian2018zero} & 37k & 50
    & & & \checkmark
    & {\footnotesize animals}\\[3pt]
   LAD \smallcite{zhao:cvprw:19}& 78k & 230
    & & & \checkmark
    & {\footnotesize animals, fruits, objects, other}\\[3pt]%
   SUN \smallcite{Patterson:cvpr:12} & 14k & 717
    & & & \checkmark
    & {\footnotesize scenes}\\[3pt]
   CUB \smallcite{Welinder:techreport:10} & 11k & 200
     & short image capts. (\dag) \smallcite{reed2016learning} & \checkmark (\ddag) \smallcite{elhoseiny2017link} & \checkmark
     & {\footnotesize birds}\\[3pt]
   NABirds \smallcite{van2015building} & 48k & 1,011
    & & \checkmark (\ddag) \smallcite{elhoseiny2017link} & \checkmark
    & \footnotesize{birds}\\[3pt]
   ImageNet-Wiki & 1.3M* & 1,000*   %
    & class names & \checkmark {\scriptsize(this work)} &
    & \footnotesize{animals, plants, objects, other}
    \\
   \bottomrule
 \end{tabular}
 }
 \end{center}
 \end{table*}

Our experimental results lead us to two significant observations.
First, models using the textual descriptions perform much better than those using class name features.
Remarkably, a simple ZSL model based on DeViSE \cite{frome2013devise} and trained on text from \NewDataset{} clearly outperforms more recent, state-of-the-art models that use word embedding features of class names as the auxiliary data.
For the CADA-VAE model~\cite{schonfeld2019generalized}, \NewDataset{} leads to almost a 12 pp. improvement (38.63\% vs. 50.50\%) in top-5 accuracy on ImageNet \textit{mp500} test split,
which is far higher than achieved in any prior work.
These results strongly suggest that algorithmic development on ZSL models is not sufficient for further substantial progress. It is necessary to also consider the quality and type of auxiliary data used. Fortunately, creating or collecting already available textual information about classes is practical, viable, and much less labor-intensive than collecting and annotating images with labels or attributes.
Our second main observation is that, regardless of the type of auxiliary data used, ZSL models generalize poorly from one large class category to another. For example, excluding all animal classes from the training set leads to large performance drops on animal species in the test set.
Therefore, the field's algorithmic progress measured on small datasets with few categories (Table~\ref{tbl:zsl_datasets}) might not be well aligned with the end goal of using ZSL to scale image classifiers to diverse sets of unseen classes from many diverse categories.  Though textual descriptions are already available for smaller datasets (CUB, NABirds) we believe \NewDataset{} will facilitate further research in this area.

\newcommand{\tabitem}{~~\llap{\textbullet}~~}

 \begin{table}[tb]\centering
   \caption{Advantages and disadvantages of different auxiliary data types for zero-shot image classification.}
   \vspace*{3pt}
{\footnotesize
\begin{tabular}{p{0.16\linewidth} p{0.75\linewidth}}
\toprule
\textbf{Auxiliary}\\
\textbf{data type} & \textbf{Advantages / Disadvantages} \\
\midrule
Attributes
&
\begin{minipage}[t]{\linewidth}
\begin{enumerate}[label=$-$,leftmargin=*, nosep,itemsep=3pt]
\item[+] \textit{Little noise}: attributes selected by experts
\item[+] \textit{Potentially highly discriminative}: depending on the selected features, might be highly informative
\item \textit{Difficult to scale}: as the number of classes increases, the more attributes are required to discriminate

\item \textit{Simple relations}: attributes may be conditional and difficult to quantify (e.g. zebra has brown and white stripes at birth)

\item \textit{Expensive annotation}: requires domain knowledge to define discriminative attributes

\item \textit{Low flexibility}: difficult to aggregate multiple independent datasets as attributes not standardized  
\end{enumerate}

\end{minipage}
\\
\\
\midrule
\begin{minipage}[t]{\linewidth}  
  Word\newline
  embedding\newline
  of class\newline names
\end{minipage}
&
\begin{minipage}[t]{\linewidth}
\begin{enumerate}[label=$-$,leftmargin=*, nosep,itemsep=3pt]
\item[+] \textit{Very simple approach}: little or no labor required to create data
\item \textit{Sensitive to linguistic issues}: names may partially overlap or not reflect well semantic similarity
\item \textit{Little information}: the name gives little discriminability of classes  
\end{enumerate}
\end{minipage}
\\
\\
\midrule
\begin{minipage}[t]{\linewidth}
  Text \newline descriptions
\end{minipage}
&
\begin{minipage}[t]{\linewidth}
\begin{enumerate}[label=+,leftmargin=*, nosep,itemsep=3pt]
\item \textit{Rich in information}: likely to be discriminative
\item \textit{Easy to collect}: many existing resources
\item \textit{Easy aggregation}: allows conceptually easy aggregation of text from multiple sources
\item[$-$] \textit{High noise}: contains non-relevant information
\end{enumerate}
\end{minipage}
\\
\bottomrule
\end{tabular}
}
\label{tbl:aux_data_pros_cons}
\end{table}

The textual descriptions for ZSL lead to significant performance improvements, but they also have several theoretical and practical advantages over other types of auxiliary data, as we detail in Table~\ref{tbl:aux_data_pros_cons}.
Moreover, we believe a large-scale zero-shot learning setup with rich and diverse textual descriptions can be a useful for studying general multimodal machine learning, specifically on the interaction between language and vision. This is because the task is challenging and requires effective modeling of the interaction between modalities for better-than-random accuracy (unlike say VQA where relying on language priors may be possible \cite{balanced_binary_vqa, balanced_vqa_v2}).
Additionally, \NewDataset{} covers a broad range of many classes, with Wikipedia articles that are long, complex, diverse, and written by many authors.
We expect that this experimental setting will lead to more progress as methods become more tailored to relating text language and vision.

\paragraph{Summary of the contributions}
We demonstrate that simply using better auxiliary data in the form of Wikipedia description of classes outperforms all prior works on zero-shot learning on the \textit{mp500} test split of ImageNet. Additionally, we show ZSL models generalize very poorly to novel super-categories of classes. Finally, we introduce \NewDataset{}, which provides text from Wikipedia articles of ImageNet classes, which we hope will facilitate the research on ZSL and the interaction between language and vision.

\section{Related works}

\paragraph{Zero-shot learning}
The problem of zero-shot learning (ZSL) was initially defined and introduced by \citet{larochelle2008zero}.
A significant fraction of the recent works on ZSL uses attributes as auxiliary data, and small or medium-sized datasets
\cite{xian2018zero,romera2015embarrassingly,changpinyo2016synthesized,schonfeld2019generalized}.
The datasets generally used contain either animal species (CUB, AWA1, AWA2), flower species (Oxford Flower Dataset), or a broader set of classes (SUN, aPY). Some of the works experiment on ImageNet, but due to the lack of attribute annotations, they use word2vec embeddings of class names instead \cite{xian2018zero, changpinyo2016synthesized, schonfeld2019generalized}.
Creating discriminative attribute annotations becomes more challenging as the number of classes grows, and the differences between classes become more nuanced.
We would like the \NewDataset{} dataset to stimulate future research on large-scale ZSL as it provides a semantically rich source of auxiliary data, shown by our experiments, for ImageNet.

\paragraph{Zero-shot learning from text}
The portion of ZSL research that is more closely related to our work uses text as auxiliary data.
\citet{reed2016learning} collected a dataset with multiple single-sentence image-level descriptions for two datasets: CUB-2011 containing bird species and Oxford Flowers dataset with flower species.
The general idea of utilizing Wikipedia articles for ZSL has already been studied by \citet{elhoseiny2016write,elhoseiny2017link} on datasets with bird species (CUB-2011, NABirds).
The usefulness of such data was more recently additionally supported by \citet{zhu2018generative} and \citet{chen2020canzsl}.
These works were, however, limited to relatively small datasets with only closely related classes compared to ImageNet, which is a much larger dataset and covers a more diverse set of classes.

\section{ImageNet--Wikipedia correspondences}
ImageNet classes correspond to a subset of synsets representing nouns in WordNet 3.0.
Each class is defined by its wnid (WordNet ID), class phrases (which we also refer to as \textit{class names} in this paper), and a gloss (brief description), for example (n02129165; ``lion, king of beasts, Panthera leo''; \textit{large gregarious predatory feline of Africa and India having a tawny coat with a shaggy mane in the male}).
The synsets have a hierarchical structure, such that our example class is a descendant
of classes like ``big cat'' and ``animal''.

We create an automatic matching of ImageNet classes to their corresponding Wikipedia pages.\footnote{We use enwiki-20200120 dump from \url{https://dumps.wikimedia.org/backup-index.html} (Accessed: 22 Jan 2020).}
The matching is based on the similarity between the synset \textit{words} and Wikipedia titles, as well as their ancestor categories.
Unfortunately, such matching occasionally produces false-positives, as it is a difficult problem.
One reason for this could be that classes with similar names might represent very different concepts -- e.g. (n02389026; ``sorrel''; \textit{a horse of a brownish orange to light brown color}) with the herb called ``sorrel''.
To ensure high-quality correspondences in \NewDataset{}, we first compared all automatic correspondences from both \citet{niemann2011people} and \citet{matuschek2013dijkstra}, then also manually verified all the matches, and modified them if necessary.
The quality of the automatic matches often depends more, for example, on how irrelevant articles are filtered out than on the exact matching approach taken. We refer the interested reader to the source code we provide for the full details.

\section{Encoding textual descriptions for zero-shot learning}
The text for each class is encoded into a single feature vector.
Having features of fixed dimensionality for each class allows us to use standard ZSL approaches.
As text encoders we use either
ALBERT~\cite{lan2019albert}, which is an NLP model based on Transformer~\cite{vaswani2017attention},
or word embeddings models: GloVe~\cite{pennington2014glove} or word2vec~\cite{mikolov2013distributed}.

For ALBERT we consider two official pre-trained models:
ALBERT-base and ALBERT-xxlarge, which have hidden layer sizes of 768 and 4096 respectively.
ALBERT~\cite{lan2019albert}, similarly to its predecessor BERT~\cite{devlin2019bert}, uses learnable positional encoding, which has fixed size.
Therefore in order to encode pages that can be longer than the maximal length, we split the text into partially overlapping sequences of 256 tokens with an overlap of 50 tokens. We encode each of the sequences with ALBERT,
and average the hidden states from the last layer over each token.
Such representations of all the sequences are then also averaged to get the final feature vector.
We compared alternative ways to extract features in Appendix~\ref{app:different_albert_encodings}.

For word embedding models, either GloVe~\cite{pennington2014glove} or word2vec~\cite{mikolov2013distributed}, we use official pre-trained weights. To encode the text we simply embed each token and then average the embedding features over the whole sequence.
Finally, if a class has multiple corresponding articles, we average the representations obtained from each of them.

\section{Experiments}
To compare the quality of different types of auxiliary data, we run experiments on two existing ZSL models.
We choose to run all experiments on the standard ZSL task, instead of generalized ZSL (GZSL).
That means we predict only unseen classes and not the union between seen and unseen classes.
Although GZLS is the more practical long-term scenario, the standard ZSL setup better isolates the usefulness and discriminativeness of auxiliary data from the model's level of overfitting to the seen classes.
As we introduce no new methods and focus on the importance of auxiliary data instead, the standards ZSL setup is more informative in our case.

\subsection{Zero-shot learning models}

We run experiments with a state-of-the-art 
CADA-VAE~\cite{schonfeld2019generalized} model.
Although it was initially proposed for GZSL, we found that it also works well for the standard ZSL setup.
To compare the relative importance of the quality of the model vs. the quality of information contained in auxiliary data, the second model we use is a simple approach based on linear projections. We refer to it as \textit{Simple ZSL}.

\paragraph{CADA-VAE~\cite{schonfeld2019generalized}}
The model is based on two Variational Auto-Encoders (VAEs)~\cite{diederik2014auto} - one for image features and another for auxiliary data.
It aligns the latent spaces of the two VAEs, such that the latent features of an image encoded with the image feature VAE and its class features encoded with auxiliary data VAE should be close to each other.
After training the VAEs,
a linear classifier is trained from the latent space vectors, obtained from the text description of the unseen classes, to the class labels of these unseen classes.\footnote{This describes the ZSL setting. Originally for GZSL, the latent features from the image samples are also used.}
At test time the latent space features obtained from the image VAE are passed to the classifier.
We use our own implementation of the model which yields similar results to those reported in the original paper.
The details can be found in Appendix~\ref{app:model_details},
along with a discussion on the small differences between our and the original implementation.

\paragraph{Simple ZSL model}
We also create a simple zero-shot learning model inspired by DeViSE~\cite{frome2013devise}.
Unlike the original work, we do not fine-tune the visual model and use two separate linear transformations for image features and auxiliary data to map the different modalities to a joint embedding space.
The transformations are trained using a hinge loss which is defined:
Assume there
are $K$ classes and $\vect{x}$ is the visual feature representation of a
sample with class label $1 \leq y \leq K$ then the loss is
\begin{align}
    \sum_{\substack{i=1 \\ i \neq y}}^K
    \max \left(0,
    m
    -
    \phi_x (\vect{x})^T \phi_t (\vect{t}_y)
    +
    \phi_x (\vect{x})^T \phi_t (\vect{t}_i)
    \right)
\end{align}
where $\vect{t}_i$ is the feature vector for the auxiliary data of class $i$, $m \geq 0$ is a scalar corresponding to the margin,
$\phi_x(\vect{x}) = W_x\,\vect{x} + \vect{b}_x$ and $\phi_t(\vect{t}_i) = W_t\,\vect{t}_i + \vect{b}_t$.
Then $W_x, W_t, \vect{b}_x$ and $\vect{b}_t$ are the trainable parameters in the model.
At test time, to classify samples, we use the nearest neighbors between projected visual features and text labels.

\subsection{Experimental setup}

Our textual descriptions are assessed by comparing them to using word2vec embeddings~\cite{mikolov2013distributed} of class names as the auxiliary data provided by \citet{changpinyo2016synthesized}.
The latter is a standard feature used in zero-shot learning on ImageNet~\cite{xian2018zero,xian2019f,schonfeld2019generalized}.
These word2vec representations were also learned on Wikipedia data.
However, this is different from encoding explicit textual descriptions of classes.

\paragraph{Data}
We use standard splits between ImageNet classes proposed by \citet{xian2018zero}.
The \textit{train} and \textit{val} splits consist of 750 and 250 classes, which contain images from ILSVRC-2012~\cite{russakovsky2015imagenet}.
Even though the original setup consists of multiple sets of separate test classes, we only use one of them: \textit{mp500} - the 500 most populated classes.
We leave studying other splits for future work since it requires providing the corresponding Wikipedia articles for the additional classes in those splits.

\paragraph{Features}

To represent images we use the 2048-dimensional features provided by \citet{xian2018zero}, which come from the last layer of a ResNet-101 before pooling. The model was pre-trained on ILSVRC 2012 part of ImageNet, which corresponds to the \textit{trainval} ZSL split we use.
Word2vec vector features of ImageNet class names (refered to as word2vec*) are from \citet{changpinyo2016synthesized} have 500-dimensions and were trained on Wikipedia data.
These are standard features used on ImageNet and were used by all the models we compare against.
For encoding the article text with word2vec, we instead use standard pre-trained 300-dimensional vectors provided by \citet{mikolov2013distributed}. GloVe features are also 300-dimensional and are from \citet{pennington2014glove}.
In all the experiments, we keep both image and auxiliary features fixed.

\paragraph{Choice of the hyper-parameters}
The hyperparameter values of the models are selected independently for each model and type of auxiliary data feature.
The models are trained on the set of  \textit{train} classes and their performance evaluated on the \textit{val} classes.
We use both random and manual searches over possible settings, and the same number of runs was made on each model variant. More details are in Appendix~\ref{app:model_details}.
The \textit{val} set consists of a subset of classes that were used to train the image feature extractor,
which violates ZSL assumptions. Although that is likely to lead to over-optimistic numbers, \textit{val} performance is used only for the hyperparameter selection.

\paragraph{Evaluation}
We evaluate the models using the mean per class top-1 and top-5 accuracy which are standard ZSL metrics.
All the hyperparameter values used for evaluation on \textit{mp500} set were solely determined based on \textit{val} performance of the models. For the final evaluation, we train the models on \textit{train}+\textit{val} with 1000 classes (750 \textit{train} + 250 \textit{val}) and use separate 500 classes from \textit{mp500} for measuring performance.
Since \NewDataset{} contains auxiliary features for 489 out of 500 classes from \textit{mp500}, unless stated differently, for a fair comparison, we compute the accuracies of the models using Wikipedia articles assuming 0 accuracy on the remaining classes.

\subsection{Results}

\paragraph{Comparison of different auxiliary data encodings}
First, in Table~\ref{tbl:val_results_aux_features}, we compare ways to encode the auxiliary data on the \textit{val} set only. We observe that using the whole Wiki articles works better than using just the abstracts (first paragraphs of the articles before any section starts). Also, word embedding encoding of the text appears to work better than the more complex ALBERT model~\cite{lan2019albert} with GloVe~\cite{pennington2014glove} being the best feature extractor.
For completeness of the comparison, we also try to encode class names with ALBERT and GloVe, although ALBERT was designed to work with longer sequences, therefore it can perform poorly with simple class names.

\begin{table}[ht]\centering
\caption{%
    \textbf{Validation set comparison of different textual inputs and encodings for zero-shot learning on ImageNet}.
    750 \textit{train} classes were used as seen and 250 \textit{val} classes as unseen. 
    Hyperparameters for each features type were tuned only on Wiki article data.
    All ALBERT encodings are of type ALBERT (xxlarge).
    word2vec* refers to the features provided by~\citet{xian2018zero} typically used for ZSL, which are different from the standard pretrained word2vec model.
}
\vspace*{3pt}
\ra{1.3}
{\small
\begin{tabular}{@{}*{2}{l}*{2}{r}@{}}
  \toprule
  & & \multicolumn{2}{c}{\textbf{Result on val}}\\
  \cmidrule{3-4}
  \textbf{Auxiliary data} & \textbf{Features} & top-1 (\%)& top-5 (\%)\\
  \midrule
  Wiki article & GloVe & {\bftab 40.24} & {\bftab 77.96}\\
  Wiki article & word2vec & 36.14 & 72.80\\
  Wiki article & ALBERT %
  & 31.29 & 64.92\\
  Wiki abstract & ALBERT %
  & 22.71 & 52.84\\
  Class names & ALBERT %
  & 14.19 & 30.53\\
  Gloss & ALBERT 
  & 19.67 & 48.09\\
  Class names \& gloss & 
  ALBERT 
  & 21.38 & 49.75\\
  Class names & GloVe
  & 29.47 & 59.26\\
  Gloss & GloVe
  & 22.41 & 54.91\\
  Class names \& gloss & GloVe
  & 33.26 & 65.47\\
  \midrule
  Class names* & word2vec* & 27.87 & 61.91\\
  \bottomrule
\end{tabular}
}
\label{tbl:val_results_aux_features}
\end{table}

\begin{table*}[ht]\centering
\caption{Zero-shot learning results of the models we evaluate on ImageNet dataset.
word2vec* uses features from \citet{changpinyo2016synthesized} and is different from word2vec which are standard pre-trained vectors.
We report the mean and standard deviation values from 5 training runs with different random seeds.
}
{\small
\begin{tabular}{@{}*{3}{l}*{2}{r}@{}}
  \toprule
  & & & \multicolumn{2}{c}{\textbf{Result on mp500}}\\
  \cmidrule{4-5}
  \textbf{Model} & \textbf{Auxiliary data} & \textbf{Features} & top-1 (\%) & top-5 (\%) \\
  \midrule
  CADA-VAE~\cite{schonfeld2019generalized} & class names & word2vec* & 15.99 $\pm$ 0.25 & 38.63 $\pm$ 0.43\\
  CADA-VAE~\cite{schonfeld2019generalized} & Wiki articles & ALBERT (base) & 17.04 $\pm$ 0.13 & 40.23 $\pm$ 0.27\\
  CADA-VAE~\cite{schonfeld2019generalized} & Wiki articles & ALBERT (xxlarge) & 18.84 $\pm$ 0.21      & 43.94 $\pm$ 0.24\\
  CADA-VAE~\cite{schonfeld2019generalized} & Wiki articles & word2vec & 20.34 $\pm$ 0.31      & 47.50 $\pm$ 0.01\\
  CADA-VAE~\cite{schonfeld2019generalized} & Wiki articles & GloVe & {\bftab 22.27 $\pm$ 0.20}      & {\bftab 50.50 $\pm$ 0.33}\\
  \midrule
  Simple ZSL & class names & word2vec* & 9.08 $\pm$ 0.18            & 27.31 $\pm$ 0.23              \\
  Simple ZSL & Wiki articles & ALBERT (base) & 13.91 $\pm$ 0.23           & 37.31 $\pm$ 0.18\\
  Simple ZSL & Wiki articles & ALBERT (xxlarge) & {\bftab 16.60 $\pm$ 0.08}      & 42.39 $\pm$ 0.30\\
  Simple ZSL & Wiki articles & word2vec & 14.84 $\pm$ 0.24      & 42.93 $\pm$ 0.20\\
  Simple ZSL & Wiki articles & GloVe & {\bftab 16.58 $\pm$ 0.19}      & {\bftab 45.26 $\pm$ 0.24}\\

\bottomrule
\end{tabular}
}
\label{tbl:test_results}
\end{table*}

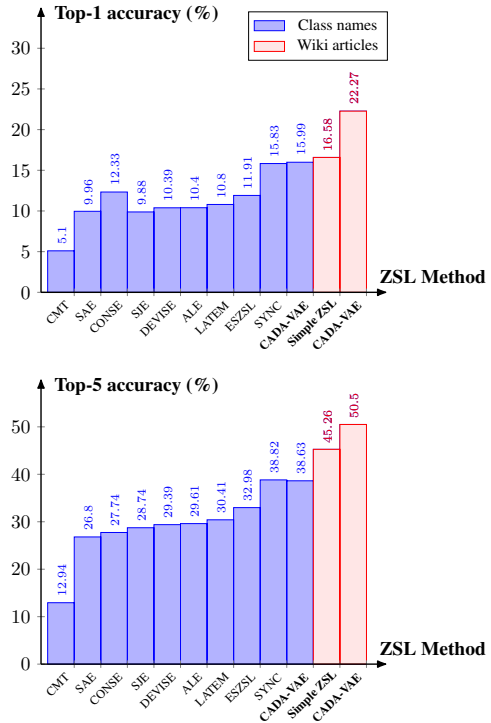
\begin{figure}[!h]\centering
  \begin{tabular}{cc}
  \resizebox{.85\columnwidth}{!}{
    \begin{tikzpicture}[>=latex]
      \begin{axis}[
          ymin=0, ymax=35,
          ytick = {0, 5, 10, 15, 20, 25, 30},
          xmin=0, xmax=12.5,
          xtick = {1, 2, 3, 4, 5, 6, 7, 8, 9, 10, 11, 12},
          xticklabels ={CMT \smallcite{Socher:nips:13}, SAE \smallcite{Kodirov:cvpr:17}, CONSE \smallcite{Norouzi:iclr:14}, SJE \smallcite{Akata:cvpr:15}, DEVISE \smallcite{frome2013devise}, ALE \smallcite{akata:tpami:16}, LATEM \smallcite{Xian:cvpr:16}, ESZSL \smallcite{romera2015embarrassingly}, SYNC \smallcite{changpinyo2016synthesized}, \textbf{CADA-VAE} \smallcite{schonfeld2019generalized}, \textbf{Simple ZSL}, \textbf{CADA-VAE} \smallcite{schonfeld2019generalized}},
    x tick label style={rotate=45, anchor=east, align=center, font=\scriptsize},
    xlabel = {\textbf{ZSL Method}},
    ylabel = {\textbf{Top-1 accuracy (\%)}},
    axis x line=bottom,
    axis y line=left,
    enlarge x limits=0.02,
    ylabel style={at={(current axis.above origin)},anchor=north west, yshift=-35pt, xshift=5pt},
    ylabel style={rotate=-90},
    xlabel style={at={(current axis.east)},anchor=east, yshift=-57pt, xshift=60pt},
    axis line style={-Latex[round],thick},
    area style,
    every node near coord/.append style={rotate=90, anchor=north west,font=\scriptsize, yshift=-2pt},
    visualization depends on={y\as\YY},
    nodes near coords={\pgfmathtruncatemacro{\YY}{ifthenelse(\YY==0,0,1)}\ifnum\YY=0\else\pgfmathprintnumber\pgfplotspointmeta\fi},
    legend style={at={(0.6,0.9)},anchor=west, column sep=5pt}
        ]
\addplot+[ybar interval,mark=no] plot coordinates {(0.0, 5.10) (1.0, 9.96) (2.0, 12.33) (3.0, 9.88) (4.0, 10.39) (5.0, 10.40) (6.0, 10.80) (7.0, 11.91) (8.0, 15.83) (9.0, 15.99) (10.0, 16.58) (11.0, 22.27) (12, 0)};
\addplot+[ybar interval,mark=no, fill=red!10] plot coordinates { (10.0, 16.58) (11.0, 22.27) (12, 0)};

      \addlegendentry{\footnotesize Class names}%
      \addlegendentry{\footnotesize Wiki articles}%
\end{axis}
  \end{tikzpicture}
  }
  \\
\resizebox{.85\columnwidth}{!}{
  \begin{tikzpicture}[>=latex]
    \begin{axis}[
    ymin=0, ymax=60,
    xmin=0, xmax=12.5,
    ytick = {0, 10, 20, 30, 40, 50},
    xtick = {1, 2, 3, 4, 5, 6, 7, 8, 9, 10, 11, 12},
    xticklabels ={CMT \smallcite{Socher:nips:13}, SAE
      \smallcite{Kodirov:cvpr:17}, CONSE \smallcite{Norouzi:iclr:14}, SJE
      \smallcite{Akata:cvpr:15}, DEVISE \smallcite{frome2013devise}, ALE
      \smallcite{akata:tpami:16}, LATEM \smallcite{Xian:cvpr:16}, ESZSL
      \smallcite{romera2015embarrassingly}, SYNC \smallcite{changpinyo2016synthesized},
      \textbf{CADA-VAE} \smallcite{schonfeld2019generalized}, \textbf{Simple ZSL}, \textbf{CADA-VAE} \smallcite{schonfeld2019generalized}},
    x tick label style={rotate=45, anchor=east, align=center, font=\scriptsize},
    xlabel = {\textbf{ZSL Method}},
    ylabel = {\textbf{Top-5 accuracy (\%)}},
    axis x line=bottom,
    axis y line=left,
    enlarge x limits=0.02,
    ylabel style={at={(current axis.above origin)},anchor=north west, yshift=-35pt, xshift=5pt},
    ylabel style={rotate=-90},
    xlabel style={at={(current axis.east)},anchor=east, yshift=-57pt, xshift=60pt},
    axis line style={-Latex[round],thick},
    area style,
    every node near coord/.append style={rotate=90, anchor=north west,font=\scriptsize, yshift=-2pt},
    visualization depends on={y\as\YY},
    nodes near coords={\pgfmathtruncatemacro{\YY}{ifthenelse(\YY==0,0,1)}\ifnum\YY=0\else\pgfmathprintnumber\pgfplotspointmeta\fi}
      ]
      \addplot+[ybar interval,mark=no] plot coordinates {(0.0, 12.94) (1.0, 26.80) (2.0, 27.74) (3.0, 28.74) (4.0, 29.39) (5.0, 29.61) (6.0, 30.41) (7.0, 32.98) (8.0, 38.82) (9.0, 38.63) (10.0, 45.26) (11.0, 50.50) (12, 0)};
      \addplot+[ybar interval,mark=no, fill=red!10] plot coordinates { (10.0, 45.26) (11.0, 50.50) (12, 0)};
\end{axis}
\end{tikzpicture}
}
  \end{tabular}
\caption{
Reported results for zero-shot learning on the ImageNet dataset and the mp500 test split.
Color indicates the auxiliary data type used.
Methods in bold-face correspond to the implementations in our paper, the other numbers were reported by~\citet{xian2018zero}.
We chose the best performing variants of our models from Table~\ref{tbl:test_results}.
}
\label{fig:zsl_test_results}
\end{figure}

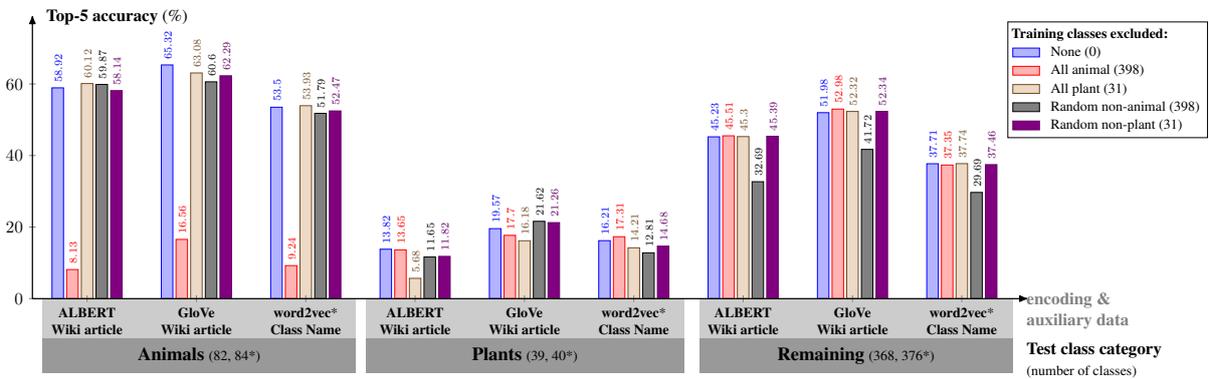
\begin{figure*}[ht]
     \def\numsize{\scriptsize}
\resizebox{\textwidth}{!}{
\begin{tikzpicture}[>=latex]
  
  \begin{axis}[
      width=24cm,
      height=8cm,
    ybar,
    ymin=0, ymax=79,
    xmin=.5, xmax = 9.6,
    enlarge x limits=0.02,    
    bar width=0.26cm,
    ylabel style={at={(current axis.above origin)},anchor=north west, yshift=-40pt, xshift=8pt, rotate=-90},
    xlabel style={at={(current axis.east)},anchor=east, yshift=-100pt, xshift=90pt,align=left},
    legend style={anchor=north west,legend columns=1, column sep=5pt,font=\footnotesize},
    legend cell align={left},
    ybar legend,        
    ylabel={\textbf{Top-5 accuracy} (\%)},
    xlabel = {\textcolor{gray}{\textbf{encoding \&}}\\ \textcolor{gray}{\textbf{auxiliary data}} \\[5pt] \textbf{Test class category} 
    \\ {\small{(number of classes)}}
    }, 
    axis x line=bottom,
    axis y line=left,    
    xtick={1,2,3, 4, 5, 6, 7, 8, 9},
    xticklabels = {\textbf{ALBERT}\\[1pt] \textbf{Wiki article}, \textbf{GloVe} \\[1pt] \textbf{Wiki article}, \textbf{word2vec}*\\[1pt] \textbf{Class Name}, \textbf{ALBERT}\\[1pt] \textbf{Wiki article}, \textbf{GloVe} \\[1pt] \textbf{Wiki article}, \textbf{word2vec}*\\[1pt] \textbf{Class Name}, \textbf{ALBERT}\\[1pt] \textbf{Wiki article}, \textbf{GloVe} \\[1pt] \textbf{Wiki article}, \textbf{word2vec}*\\[1pt] \textbf{Class Name}},
    x tick label style={font=\footnotesize, align=center},        
    axis line style={-Latex[round],thick},
    area style,
    every node near coord/.append style={rotate=90, anchor=north west,font=\scriptsize, yshift=5.5pt, xshift=0pt},
    visualization depends on={y\as\YY},
    nodes near coords={\pgfmathtruncatemacro{\YY}{ifthenelse(\YY==0,0,1)}\ifnum\YY=0\else\pgfmathprintnumber\pgfplotspointmeta\fi}        
    ]

    \begin{scope}[on background layer]
      \fill[black!20,opacity=1] ({rel axis cs:.01,-.14}) rectangle ({rel axis cs:0.325,0});
      \fill[black!20,opacity=1] ({rel axis cs:.335,-.14}) rectangle ({rel axis cs:0.655,0});
      \fill[black!20,opacity=1] ({rel axis cs:.67,-.14}) rectangle ({rel axis cs:0.985,0});      
      \fill[black!40,opacity=1] ({rel axis cs:.01,-.26}) rectangle
      node[black]{\large \textbf{Animals} {\small (82, 84*)}} ({rel axis cs:0.325,-.14});
      \fill[black!40,opacity=1] ({rel axis cs:.335,-.26}) rectangle node[black]{\textbf{\large Plants} {\small (39, 40*)}} ({rel axis cs:0.655,-.14});
      \fill[black!40,opacity=1] ({rel axis cs:.67,-.26}) rectangle node[black]{\textbf{\large Remaining} {\small (368, 376*)}} ({rel axis cs:0.985,-.14});            
    \end{scope}
    
  \addlegendimage{empty legend}

\addplot coordinates {(1,58.92) (2,65.32) (3,53.50) (4,13.82) (5,19.57) (6, 16.21) (7,45.23) (8,51.98) (9,37.71)};  
\addplot coordinates {(1,8.13) (2,16.56) (3,9.24)  (4,13.65) (5,17.70) (6,17.31) (7,45.51)  (8,52.98) (9, 37.35)};
\addplot coordinates {(1,60.12) (2,63.08) (3,53.93) (4,5.68) (5,16.18) (6,14.21) (7,45.30) (8,52.32) (9,37.74)};
\addplot coordinates {(1,59.87) (2,60.60) (3,51.79) (4,11.65) (5,21.62) (6,12.81) (7,32.69)   (8,41.72)   (9,29.69)};
\addplot coordinates {(1,58.14) (2,62.29) (3, 52.47) (4,11.82) (5,21.26) (6,14.68) (7,45.39)  (8,52.34) (9, 37.46)};  

\addlegendentry{\hspace{-25pt}\textbf{\footnotesize Training classes excluded:}}
\addlegendentry{None (0)}
\addlegendentry{All animal (398)}
\addlegendentry{All plant (31)}
\addlegendentry{Random non-animal (398)}
\addlegendentry{Random non-plant (31)}
\end{axis}
\end{tikzpicture} 
}
\caption[]{\textbf{Within and across category generalization of ZSL. Top-5 per class accuracy on CADA-VAE.}
   We show the effect of excluding a category of classes from
   the training set on ZSL performance for all mp500 test set categories. Specifically we compare excluding all \textit{animal} classes vs. excluding the
  same number of random \textit{non-animal} classes. We repeat the same process
  for the \textit{plant} class.
  Different models trained on different auxiliary data are compared.
  The performance numbers are from a single run each. The performance on the unseen \textit{animal} classes drops
  dramatically when the animal classes are removed from the training set, the red bars in the first grouping of plots. The same trend, though not as pronounced, can be seen for the \textit{remaining} classes, the gray bars in the last grouping of plots.
  The number of test classes is slightly higher for models using class name data (marked with *) since Wiki articles are missing for some classes.
  }
  \label{fig:exclude_groups}
\end{figure*}

\paragraph{Evaluating models and feature types on the test set}
In Table~\ref{tbl:test_results}, we compare zero-shot learning accuracies of various models on the \textit{mp500} test split. Our experiments compare two different types of auxiliary information, class names encoded with word2vec, which is a standard approach on ImageNet, as well as using Wikipedia articles that we extract, encoded with two variants of ALBERT and two variants of word embedding encodings.
The results show that models using the textual descriptions consistently achieve higher accuracies. The increased performance is even more prominent on the simple ZSL model. When using textual descriptions, such a simple model outperforms much more complex and generally better CADA-VAE trained with word2vec features of class names.
This demonstrates not only the high quality of information the Wikipedia articles convey about the classes but also that the information can be effectively extracted even by simple models.
In Figure~\ref{fig:zsl_test_results}, we additionally compare the results we achieve with those reported by \citet{xian2018zero} of different zero-shot learning methods. The general setup we used is the same and uses the same image features, data splits and word2vec vectors of the class names.
The comparison demonstrates that the models using Wikipedia articles outperform all the prior methods, highlighting the relative importance of the auxiliary data type and algorithmic development.

\paragraph{Generalization outside the seen groups of classes}

The hierarchical structure of WordNet entities, used as ImageNet classes, allows us to study different groups of classes separately.
In Figure~\ref{fig:exclude_groups}, we split the set of both train and test classes into three disjoint partitions: \textit{animals}, \textit{plants}, and \textit{remaining} and compare the test performance on all subgroups when some of them are excluded from the training set.
First of all, the performance on \textit{plants} is generally much lower than the much more populated \textit{animals} group.
Moreover, we see that all the models have poor accuracy on \textit{animals} when that group is excluded from the training set.
These results show the models' inability to generalize to unseen classes that are semantically far from seen classes.
Therefore, they also suggest that studying ZSL on ImageNet might be more practical than datasets with a less diverse set of classes.

\begin{table*}[h!]     \centering
\caption{\textbf{
Performance on subsets of mp500 that excludes classes that are similar to those in the training set.
}
We take two CADA-VAE models (single runs), trained to predict all unseen classes and present their mean per-class accuracies of different subsets of the \textit{mp500} classes.
Test classes are excluded based on their overlap with the train+val classes in either: Wikipedia pages correspondences or WordNet phrases (defining the classes). Each row corresponds to a different subset of excluded classes. Examples of excluded classes are:
\textbf{Row 3}: (n03792972, \enquote{mountain tent}) is removed as train+val contains (n02769963, \enquote{backpacking tent}, \enquote{pack tent}) and all are mapped to \enquote{Tent}.
\textbf{Row 4}: (n04543772, \enquote{wagon wheel}) is removed as it maps to [\enquote{Wagon}, \enquote{Wheel}], and class (n02974003, \enquote{car wheel}) in train+val maps to \enquote{Wheel}.
\textbf{Row 5}: (n03222516, \enquote{doorbell, bell, buzzer}) is removed as there is (n03017168, \enquote{chime, bell, gong}) in train+val. The numbers shown in bold display the results with the biggest drop in performance.
}
\ra{1.3}
{\small
\begin{tabular}{@{}lc*{6}{r}@{}}
  \toprule
  & & \multicolumn{2}{c}{\makecell{ALBERT-xxlarge\\Wiki articles}} & \multicolumn{2}{c}{\makecell{GloVe\\Wiki articles}} & \multicolumn{2}{c}{\makecell{word2vec\\class names}}\\
  \cmidrule(l){3-4} \cmidrule(l){5-6} \cmidrule(l){7-8}
  \textbf{Classes excluded} & \textbf{\# classes left} & top-1 (\%) & top-5 (\%) & top-1 (\%) & top-5 (\%) & top-1 (\%) & top-5 (\%)\\
  \midrule
None                 & 500                  & 18.73 & 44.03     & 22.26 & 50.49   & 16.38 & 39.07              \\
No matching articles & 489                  & 19.16 & 45.02     & 22.76 & 51.63   & 16.28 & 39.00              \\
\midrule
Same set of articles matched (*) & 459      & 16.82 & 42.63     & 21.08 & 49.82   & 16.18 & 39.04              \\
(Any) article overlap            & 439      & {\bftab 15.07} & {\bftab 40.64}  & {\bftab 19.87} & {\bftab 48.38}  & 15.68 & 38.60              \\ \midrule
Class names overlap ($\dagger$) & 463   & 18.24 & 44.50     & 21.99 & 51.31   & {\bftab 15.53} & {\bftab 38.05}         \\ \midrule
(*) $\cup$ ($\dagger$)              & 438   & 16.39 & 42.42     & 20.79 & 49.76   & 15.70 & 38.35              \\
\bottomrule
\end{tabular}
}
\label{tbl:exclude_overlap}
\end{table*}

\paragraph{Analyzing the impact of class similarity}

Even though ImageNet classes represent distinct concepts, since WordNet and Wikipedia were created independently, naturally, the granularity of the concepts they define can be different.
As a consequence, some of the classes can be mapped to multiple articles.
For example, class (n02965783, ``car mirror'') was assigned two more detailed articles: ``Rear-view mirror'', ``Wing mirror''.
In fact, WordNet synsets used as ImageNet classes consist of one or more (closely related) concepts that are considered equivalent.
On the other hand, two different classes can be assigned the same set of articles.
Potential overlap of sources of auxiliary information (Wiki pages or class phrases) can be present due to a high degree of similarity between corresponding classes, or compound classes being defined on other classes (e.g. \textit{wagon wheel} being a \textit{wheel} of a \textit{wagon}).
This phenomenon is associated not only with the particular auxiliary data we propose. As we observe in Table~\ref{tbl:exclude_overlap}, models using both sources of data have lower performance on subsets of test classes which exclude those that are similar to the training classes under various criteria defined for those types of auxiliary data.
As shown in the last row, the model using Wikipedia text still outperforms the one using class phrases on the subset of test classes that exclude those that are similar to training classes in either Wikipedia pages or class phrases.
However, as demonstrated in our results in Figure~\ref{fig:exclude_groups}, certain degree of similarity between seen and unseen classes is necessary for reasonable performance.

\def\cmwidth{1.0\linewidth}
\def\wnidpathr{./imgs/figures/confusion_matrices/n03015254}

\begin{figure}[!h]\centering

    \begin{subfigure}[b]{\cmwidth}
    \includegraphics[width=\textwidth]{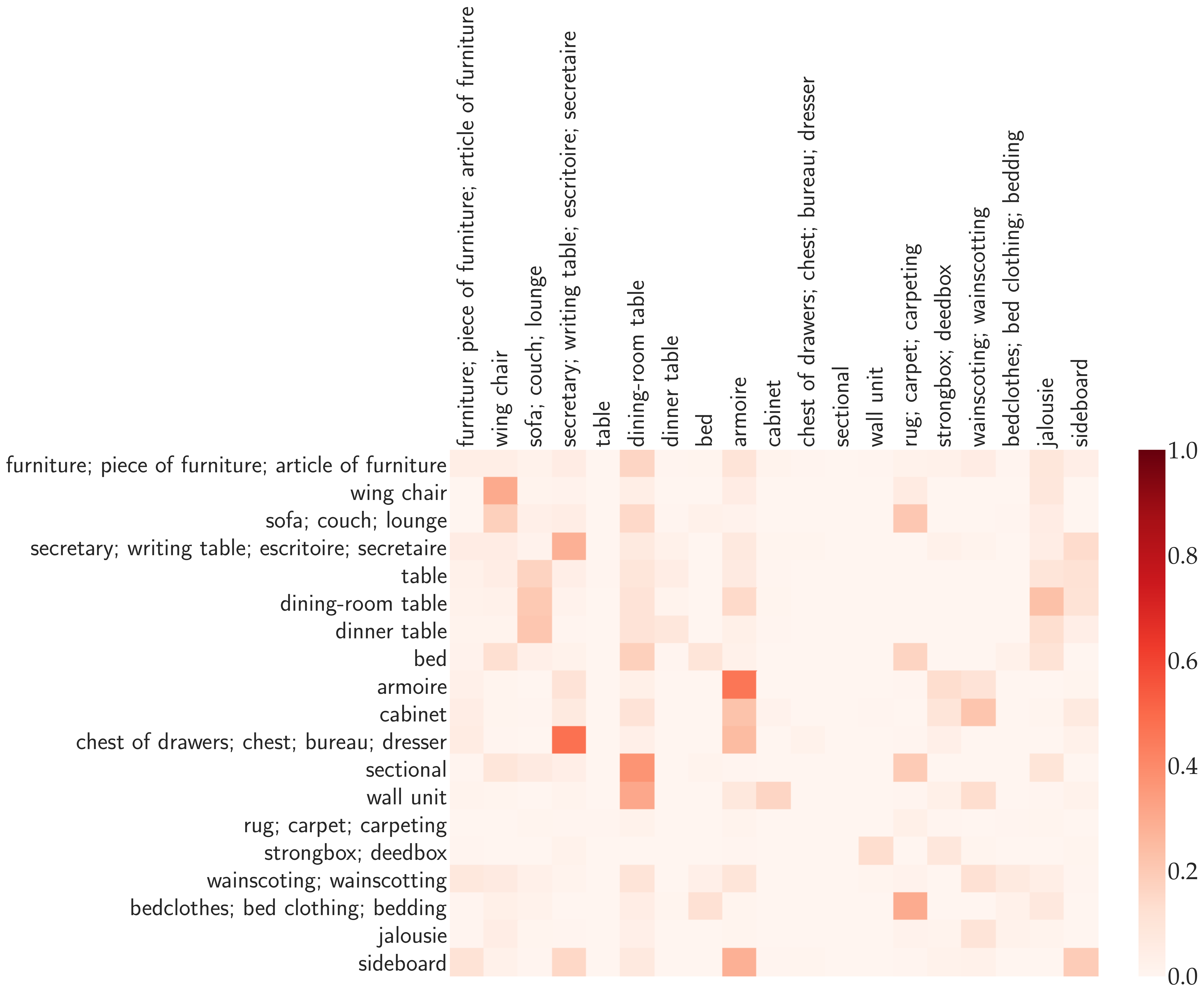}
    \caption{word2vec - class names}
    \end{subfigure}

    \begin{subfigure}[b]{\cmwidth}
    \includegraphics[width=\textwidth]{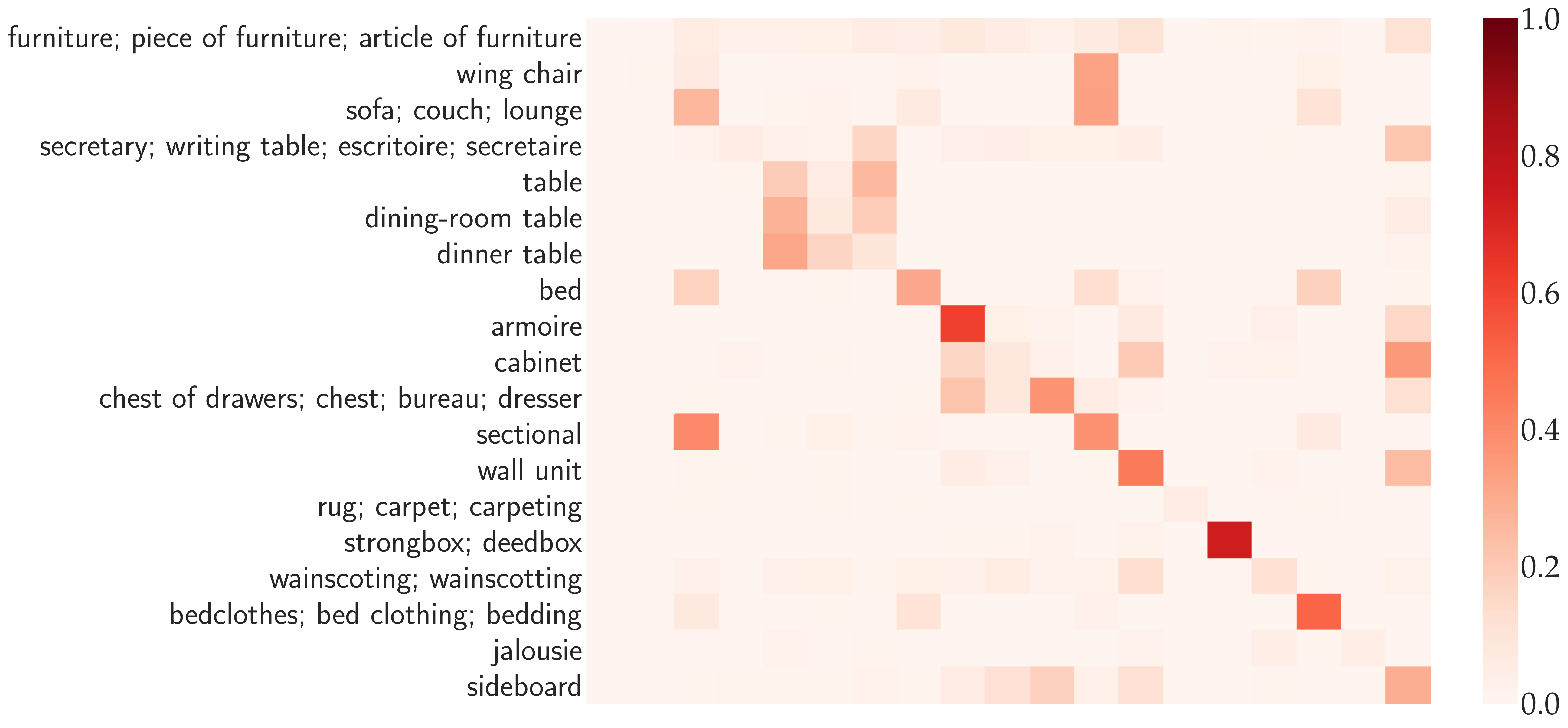}
    \caption{GloVe - Wiki articles}
    \end{subfigure}

    \begin{subfigure}[b]{\cmwidth}
    \includegraphics[width=\textwidth]{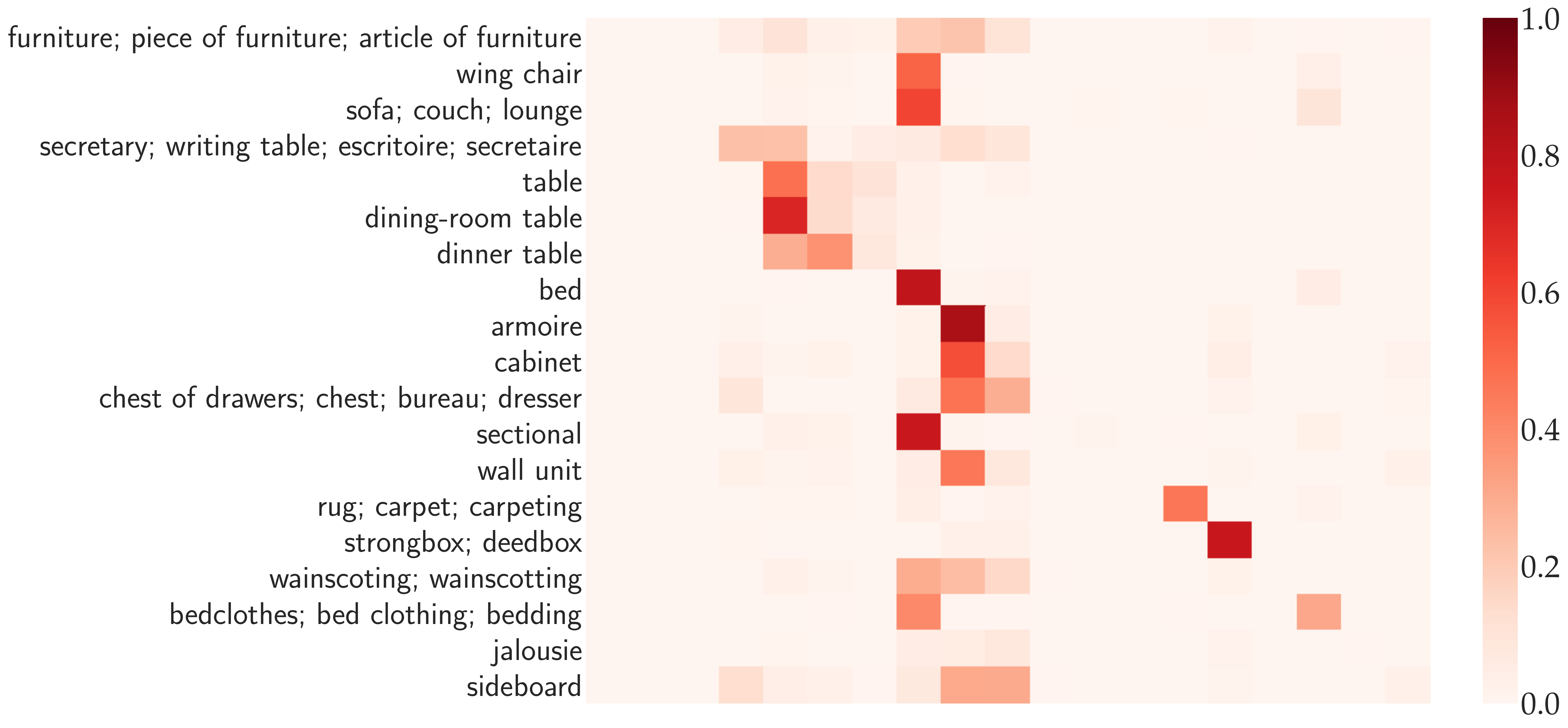}
    \caption{ALBERT-xxlarge - Wiki articles}
    \end{subfigure}

\caption{
\textbf{A selected subset of confusion matrices on mp500 test set.}
All the models are CADA-VAE~\cite{schonfeld2019generalized} trained on different types of auxiliary data.
The rows correspond to true labels, columns to predicted classes - the matrices are row-normalized.
The subset of classes was selected by choosing one class and including others that were the most frequently mutually confused by any of the models.
}
\label{fig:qualitative_confusion_matrices}
\end{figure}

\paragraph{Qualitative evaluation of prediction}

To provide further insight into models' performance, we analyze confusion matrices with predictions of different models.
Due to the immense size, the full confusion matrices are included only in the supplementary material.
Figure~\ref{fig:qualitative_confusion_matrices} shows only a selected subset of the confusion matrices, only with classes that were often confused.
The mispredictions are mainly within relatively semantically similar classes -- different types of furniture in this case.
Generally, the pattern for models using Wiki articles seems more diagonal, with more values close to 0.
This suggests that Wiki articles are richer in discriminative information.
More figures are available in Appendix~\ref{app:confusion_matrices}.

\paragraph{Text length vs. performance}
Finally, we also analyse the impact of the length of textual descriptions of classes and performance on them.
Longer descriptions tend to lead to higher accuracy. The trend is not strong, however.
The details are discussed in Appendix~\ref{app:text_length}.

\section{Limitations \& future work}
The performance of our approach is likely to depend on some general quality of Wikipedia articles. Although we have not evaluated this dependence directly in our work, 
at least the article's length indeed correlates with the corresponding class's accuracy (Appendix~\ref{app:text_length}).
Therefore, the adaptability of the general approach to some particular domains not well-covered in Wikipedia can also be limited.
Additionally, we assume that the classes used are possible to match with Wikipedia articles. With very few exceptions, this is true for the ImageNet classes we consider. However, it is possible to find cases where this would not be possible due to the different granularities of the classes and Wikipedia pages.

The ZSL methods we use can utilize the information from our textual descriptions effectively, however, they only rely on features from pretrained, off-the-shelf models. Given that the textual descriptions are complex and contain only a sparse amount of visually relevant information, future research is needed to find methods more effective at fully utilizing our textual descriptions.
Additionally, there is a potential of using our textual descriptions for different purposes, such as pretraining of image recognition models, similarly to the recent works by \citet{desai2020virtex} and \citet{radford2021learning}.
Finally, the issue of generalization to groups of classes remains an open challenge.

\section{Conclusions}
We demonstrate that simply employing Wikipedia articles as auxiliary data for ZSL on ImageNet yields a state-of-the-art performance, even on a relatively weak model. Additionally, we show that standard ZSL models poorly generalize across categories of classes. Finally, we share the \NewDataset{} dataset with Wikipedia articles describing corresponding ImageNet classes to facilitate further the research on ZSL, as well as on the interaction between language and vision more generally.

\section*{Acknowledgments}
This work was supported by the Wallenberg AI, Autonomous Systems and Software Program (WASP) funded by the Knut and Alice Wallenberg Foundation.

\bibliography{anthology,egbib}

\begin{thebibliography}{25}
\expandafter\ifx\csname natexlab\endcsname\relax\def\natexlab#1{#1}\fi

\bibitem[{Changpinyo et~al.(2016)Changpinyo, Chao, Gong, and
  Sha}]{changpinyo2016synthesized}
Soravit Changpinyo, Wei-Lun Chao, Boqing Gong, and Fei Sha. 2016.
\newblock Synthesized classifiers for zero-shot learning.
\newblock In \emph{Proceedings of the Conference on Computer Vision and Pattern
  Recognition (CVPR)}.

\bibitem[{Chen et~al.(2020)Chen, Li, Luo, Huang, and Yang}]{chen2020canzsl}
Zhi Chen, Jingjing Li, Yadan Luo, Zi~Huang, and Yang Yang. 2020.
\newblock Canzsl: Cycle-consistent adversarial networks for zero-shot learning
  from natural language.
\newblock In \emph{Proceedings of the IEEE/CVF Winter Conference on
  Applications of Computer Vision}, pages 874--883.

\bibitem[{Desai and Johnson(2020)}]{desai2020virtex}
Karan Desai and Justin Johnson. 2020.
\newblock Virtex: Learning visual representations from textual annotations.
\newblock \emph{arXiv preprint arXiv:2006.06666}.

\bibitem[{Devlin et~al.(2019)Devlin, Chang, Lee, and
  Toutanova}]{devlin2019bert}
Jacob Devlin, Ming-Wei Chang, Kenton Lee, and Kristina Toutanova. 2019.
\newblock Bert: Pre-training of deep bidirectional transformers for language
  understanding.
\newblock In \emph{Proceedings of the 2019 Conference of the North American
  Chapter of the Association for Computational Linguistics: Human Language
  Technologies, Volume 1 (Long and Short Papers)}, pages 4171--4186.

\bibitem[{Diederik et~al.(2014)Diederik, Welling et~al.}]{diederik2014auto}
P~Kingma Diederik, Max Welling, et~al. 2014.
\newblock Auto-encoding variational bayes.
\newblock In \emph{Proceedings of the International Conference on Learning
  Representations (ICLR)}, volume~1.

\bibitem[{Elhoseiny et~al.(2016)Elhoseiny, Elgammal, and
  Saleh}]{elhoseiny2016write}
Mohamed Elhoseiny, Ahmed Elgammal, and Babak Saleh. 2016.
\newblock Write a classifier: Predicting visual classifiers from unstructured
  text.
\newblock \emph{IEEE Transactions on Pattern Analysis and Machine Intelligence
  (TPAMI)}, 39(12):2539--2553.

\bibitem[{Elhoseiny et~al.(2017)Elhoseiny, Zhu, Zhang, and
  Elgammal}]{elhoseiny2017link}
Mohamed Elhoseiny, Yizhe Zhu, Han Zhang, and Ahmed Elgammal. 2017.
\newblock Link the head to the" beak": Zero shot learning from noisy text
  description at part precision.
\newblock In \emph{Proceedings of the Conference on Computer Vision and Pattern
  Recognition (CVPR)}.

\bibitem[{Frome et~al.(2013)Frome, Corrado, Shlens, Bengio, Dean, Ranzato, and
  Mikolov}]{frome2013devise}
Andrea Frome, Greg~S Corrado, Jon Shlens, Samy Bengio, Jeff Dean, Marc'Aurelio
  Ranzato, and Tomas Mikolov. 2013.
\newblock Devise: A deep visual-semantic embedding model.
\newblock In \emph{Advances in Neural Information Processing Systems (NIPS)}.

\bibitem[{Goyal et~al.(2017)Goyal, Khot, Summers{-}Stay, Batra, and
  Parikh}]{balanced_vqa_v2}
Yash Goyal, Tejas Khot, Douglas Summers{-}Stay, Dhruv Batra, and Devi Parikh.
  2017.
\newblock Making the {V} in {VQA} matter: Elevating the role of image
  understanding in {V}isual {Q}uestion {A}nswering.
\newblock In \emph{Conference on Computer Vision and Pattern Recognition
  (CVPR)}.

\bibitem[{Lan et~al.(2019)Lan, Chen, Goodman, Gimpel, Sharma, and
  Soricut}]{lan2019albert}
Zhenzhong Lan, Mingda Chen, Sebastian Goodman, Kevin Gimpel, Piyush Sharma, and
  Radu Soricut. 2019.
\newblock Albert: A lite bert for self-supervised learning of language
  representations.
\newblock In \emph{International Conference on Learning Representations}.

\bibitem[{Larochelle et~al.(2008)Larochelle, Erhan, and
  Bengio}]{larochelle2008zero}
Hugo Larochelle, Dumitru Erhan, and Yoshua Bengio. 2008.
\newblock Zero-data learning of new tasks.
\newblock In \emph{Proceedings of the 23rd national conference on Artificial
  intelligence-Volume 2}, pages 646--651.

\bibitem[{Matuschek and Gurevych(2013)}]{matuschek2013dijkstra}
Michael Matuschek and Iryna Gurevych. 2013.
\newblock Dijkstra-wsa: A graph-based approach to word sense alignment.
\newblock \emph{Transactions of the Association for Computational Linguistics},
  1:151--164.

\bibitem[{Mikolov et~al.(2013)Mikolov, Sutskever, Chen, Corrado, and
  Dean}]{mikolov2013distributed}
Tomas Mikolov, Ilya Sutskever, Kai Chen, Greg~S Corrado, and Jeff Dean. 2013.
\newblock Distributed representations of words and phrases and their
  compositionality.
\newblock In \emph{Advances in neural information processing systems}, pages
  3111--3119.

\bibitem[{Niemann and Gurevych(2011)}]{niemann2011people}
Elisabeth Niemann and Iryna Gurevych. 2011.
\newblock The people's web meets linguistic knowledge: automatic sense
  alignment of wikipedia and wordnet.
\newblock In \emph{Proceedings of the Ninth International Conference on
  Computational Semantics}, pages 205--214. Association for Computational
  Linguistics.

\bibitem[{Pennington et~al.(2014)Pennington, Socher, and
  Manning}]{pennington2014glove}
Jeffrey Pennington, Richard Socher, and Christopher~D Manning. 2014.
\newblock Glove: Global vectors for word representation.
\newblock In \emph{Proceedings of the 2014 conference on empirical methods in
  natural language processing (EMNLP)}, pages 1532--1543.

\bibitem[{Radford et~al.(2021)Radford, Kim, Hallacy, Ramesh, Goh, Agarwal,
  Sastry, Askell, Mishkin, Clark et~al.}]{radford2021learning}
Alec Radford, Jong~Wook Kim, Chris Hallacy, Aditya Ramesh, Gabriel Goh,
  Sandhini Agarwal, Girish Sastry, Amanda Askell, Pamela Mishkin, Jack Clark,
  et~al. 2021.
\newblock \href {https://openai.com/blog/clip/} {Learning transferable visual
  models from natural language supervision}.

\bibitem[{Reed et~al.(2016)Reed, Akata, Lee, and Schiele}]{reed2016learning}
Scott Reed, Zeynep Akata, Honglak Lee, and Bernt Schiele. 2016.
\newblock Learning deep representations of fine-grained visual descriptions.
\newblock In \emph{Proceedings of the IEEE Conference on Computer Vision and
  Pattern Recognition}, pages 49--58.

\bibitem[{Romera-Paredes and Torr(2015)}]{romera2015embarrassingly}
Bernardino Romera-Paredes and Philip Torr. 2015.
\newblock An embarrassingly simple approach to zero-shot learning.
\newblock In \emph{Proceedings of the International Conference on Machine
  Learning (ICML)}.

\bibitem[{Russakovsky et~al.(2015)Russakovsky, Deng, Su, Krause, Satheesh, Ma,
  Huang, Karpathy, Khosla, Bernstein et~al.}]{russakovsky2015imagenet}
Olga Russakovsky, Jia Deng, Hao Su, Jonathan Krause, Sanjeev Satheesh, Sean Ma,
  Zhiheng Huang, Andrej Karpathy, Aditya Khosla, Michael Bernstein, et~al.
  2015.
\newblock Imagenet large scale visual recognition challenge.
\newblock \emph{International journal of computer vision}, 115(3):211--252.

\bibitem[{Schonfeld et~al.(2019)Schonfeld, Ebrahimi, Sinha, Darrell, and
  Akata}]{schonfeld2019generalized}
Edgar Schonfeld, Sayna Ebrahimi, Samarth Sinha, Trevor Darrell, and Zeynep
  Akata. 2019.
\newblock Generalized zero-and few-shot learning via aligned variational
  autoencoders.
\newblock In \emph{Proceedings of the IEEE Conference on Computer Vision and
  Pattern Recognition}, pages 8247--8255.

\bibitem[{Vaswani et~al.(2017)Vaswani, Shazeer, Parmar, Uszkoreit, Jones,
  Gomez, Kaiser, and Polosukhin}]{vaswani2017attention}
Ashish Vaswani, Noam Shazeer, Niki Parmar, Jakob Uszkoreit, Llion Jones,
  Aidan~N Gomez, {\L}ukasz Kaiser, and Illia Polosukhin. 2017.
\newblock Attention is all you need.
\newblock In \emph{Advances in neural information processing systems}, pages
  5998--6008.

\bibitem[{Xian et~al.(2018)Xian, Lampert, Schiele, and Akata}]{xian2018zero}
Yongqin Xian, Christoph~H Lampert, Bernt Schiele, and Zeynep Akata. 2018.
\newblock Zero-shot learning—a comprehensive evaluation of the good, the bad
  and the ugly.
\newblock \emph{IEEE transactions on pattern analysis and machine
  intelligence}, 41(9):2251--2265.

\bibitem[{Xian et~al.(2019)Xian, Sharma, Schiele, and Akata}]{xian2019f}
Yongqin Xian, Saurabh Sharma, Bernt Schiele, and Zeynep Akata. 2019.
\newblock f-vaegan-d2: A feature generating framework for any-shot learning.
\newblock In \emph{Proceedings of the IEEE Conference on Computer Vision and
  Pattern Recognition}, pages 10275--10284.

\bibitem[{Zhang et~al.(2016)Zhang, Goyal, Summers{-}Stay, Batra, and
  Parikh}]{balanced_binary_vqa}
Peng Zhang, Yash Goyal, Douglas Summers{-}Stay, Dhruv Batra, and Devi Parikh.
  2016.
\newblock {Y}in and {Y}ang: Balancing and answering binary visual questions.
\newblock In \emph{Conference on Computer Vision and Pattern Recognition
  (CVPR)}.

\bibitem[{Zhu et~al.(2018)Zhu, Elhoseiny, Liu, Peng, and
  Elgammal}]{zhu2018generative}
Yizhe Zhu, Mohamed Elhoseiny, Bingchen Liu, Xi~Peng, and Ahmed Elgammal. 2018.
\newblock A generative adversarial approach for zero-shot learning from noisy
  texts.
\newblock In \emph{Proceedings of the IEEE conference on computer vision and
  pattern recognition}, pages 1004--1013.

\end{thebibliography}
\bibliographystyle{acl_natbib}

\appendix

\def\imgwidth{0.47\linewidth}

\begin{figure*}[!b]
\begin{tabular}{cc}
    \centering
    \includegraphics[width=\imgwidth]{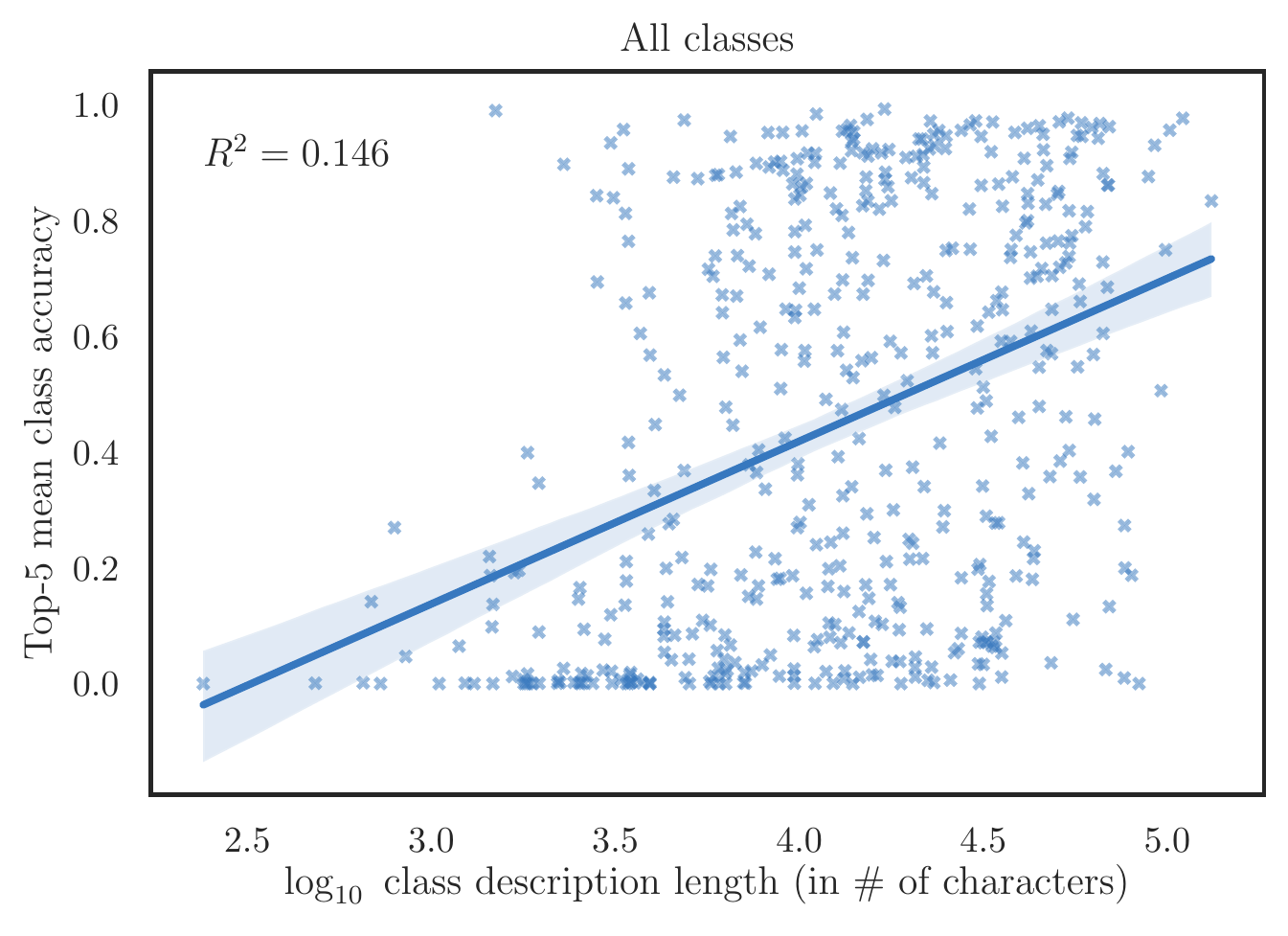}
    &
    \includegraphics[width=\imgwidth]{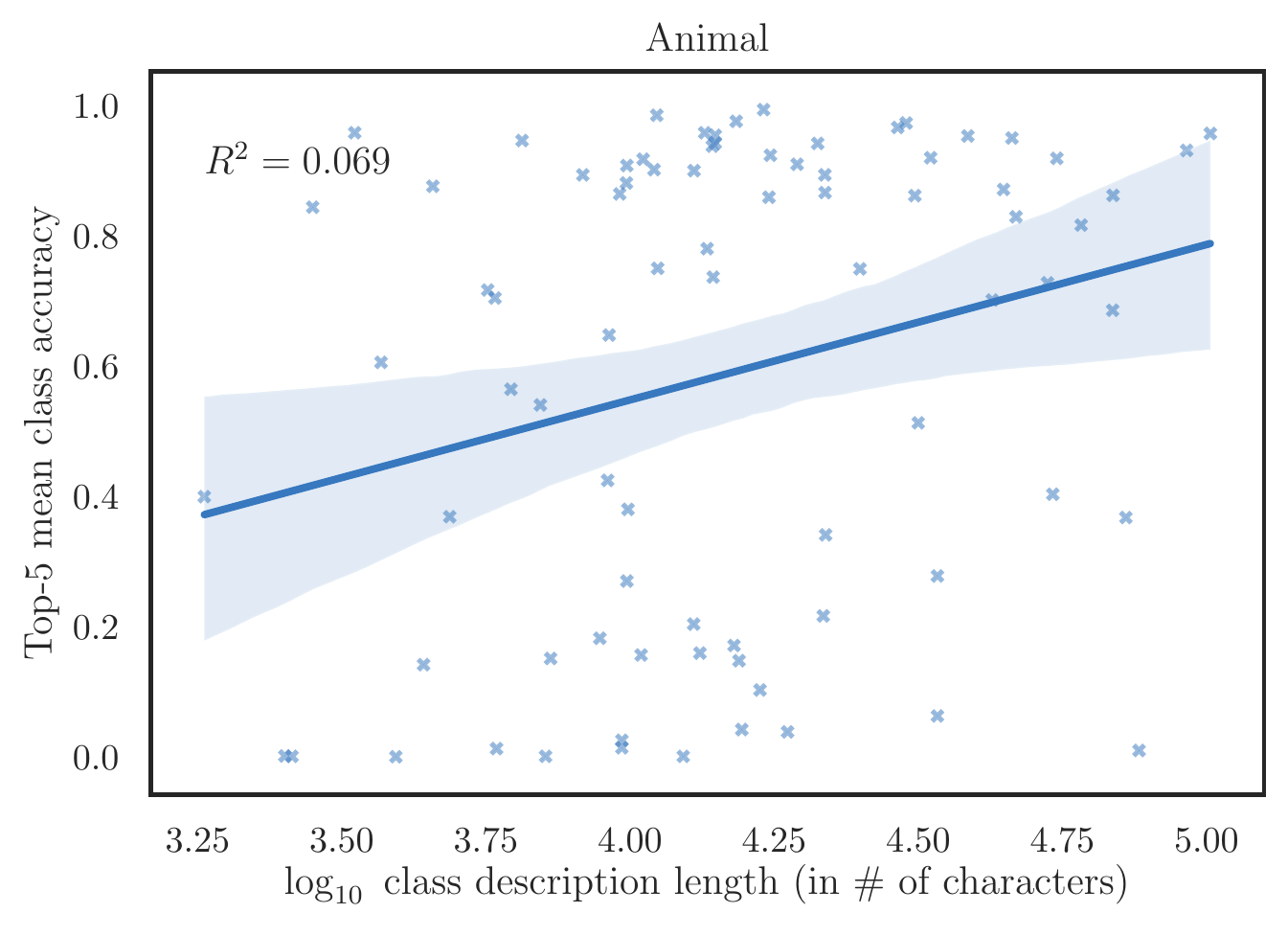} \\
    \includegraphics[width=\imgwidth]{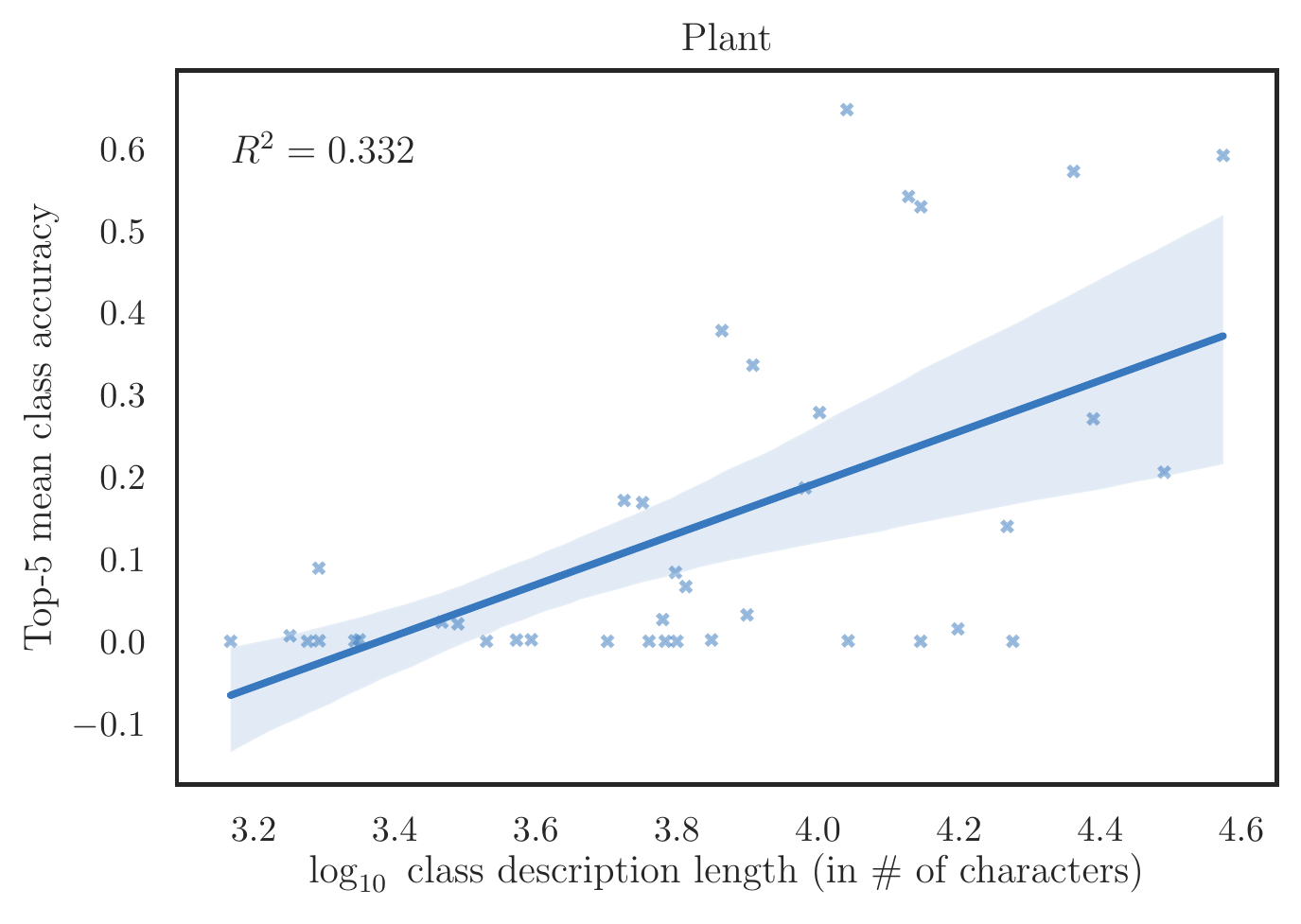}
    &
    \includegraphics[width=\imgwidth]{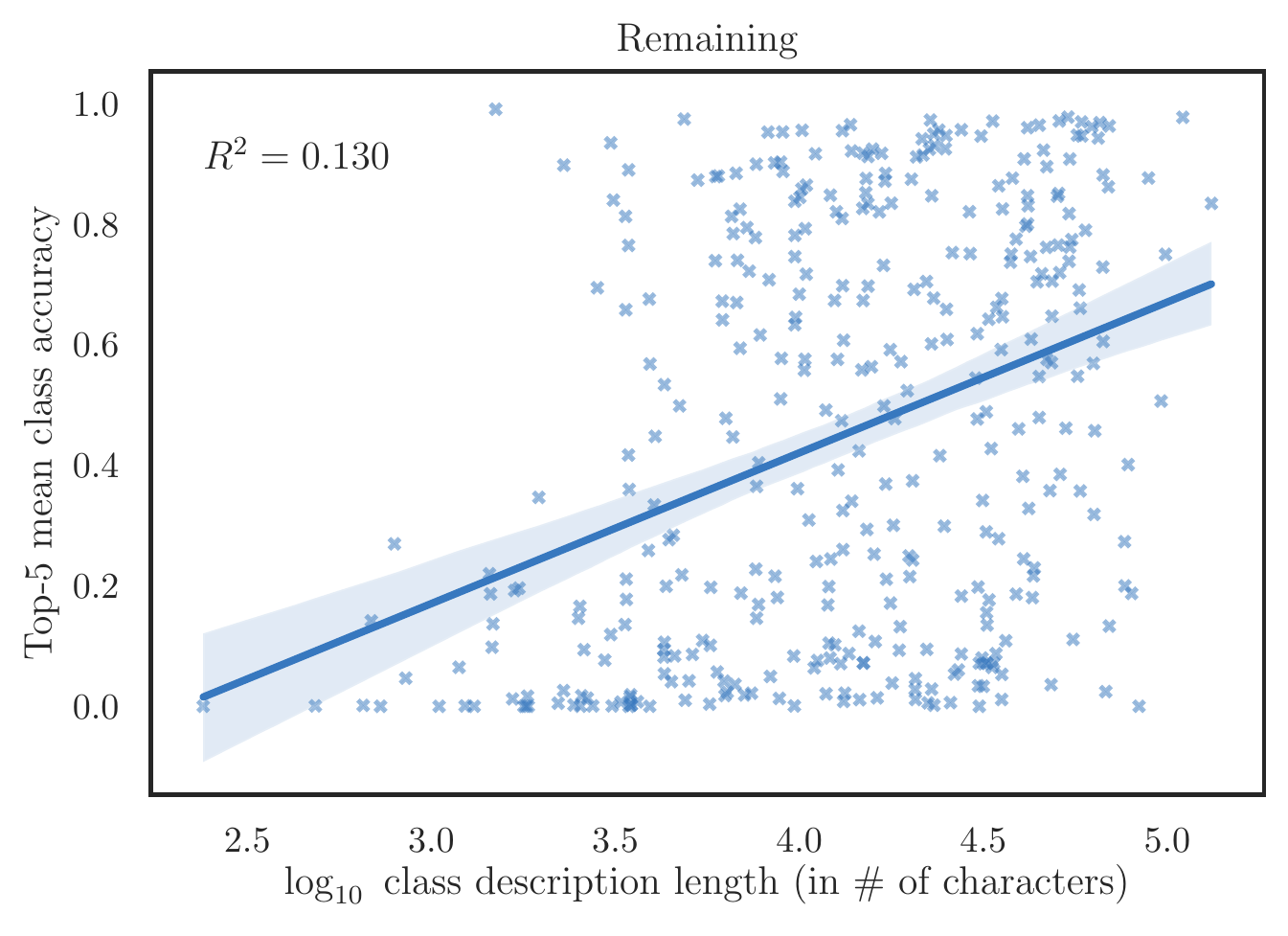}
\end{tabular}
\caption{
\textbf{The relation between the top-5 accuracy on a given class and the length of its Wikipedia pages, separately for different groups of classes from \textit{mp500} test set.}
The model used is CADA-VAE trained on the Wikipedia articles using ALBERT-xxlarge features.
The lengths are expressed in the $\log_{10}$ of the number of characters. In case a class has more than one page, we sum the lengths of all pages.
Note different ranges of axes.
}
\label{fig:text_length_vs_acc}
\end{figure*}

\section{Broader Impact}
We anticipate that the dataset we share will make research on multimodal machine learning more accessible, especially to research communities with fewer resources available.
More speculatively, we hope that in the future, work on incorporating information from multiple modalities could potentially contribute towards progress on the robustness and flexibility of machine learning models.

We would also like to point out the potential risk of the models reinforcing biases present in sources of text, in particular, in the data we use.
The authors of Wikipedia generally represent a very non-diverse social group.\footnote{\url{https://en.wikipedia.org/wiki/Gender_bias_on_Wikipedia} (Accessed: 16 Nov 2020).}
Moreover, the content of the database has been shown to contain biases, including the selection of topics it covers.\footnote{\url{https://en.wikipedia.org/wiki/Wikipedia\#Coverage_of_topics_and_systemic_bias} (Accessed: 16 Nov 2020).}
This challenge could potentially be mitigated to some degree with more research progress on interpretable machine learning and bias reduction methods.

\section{The effect of text length on accuracy}
\label{app:text_length}
In Figure~\ref{fig:text_length_vs_acc}, we show the relation between the top-5 class accuracy and the length of the textual description of that class.
In general, the classes with longer corresponding text tend to yield better accuracy, although the general trend is not strong.
However, the impact of the text length appears to be different among different class groups.
On animal classes, the relation is negligible. 
However, on plant classes the relation is much stronger.
This observation is especially important since the plant classes generally compose a relatively small fraction of seen classes, and the model has relatively low performance on this group (see Figure~\ref{fig:exclude_groups}). This result indicates that using longer textual descriptions can partially mitigate the effect of a group of classes not being well represented in the seen classes.

\def\cmwidth{0.45\linewidth}
\def\wnidpathl{./appendices/figures/confusion_matrices/n02210427}
\def\wnidpathr{./appendices/figures/confusion_matrices/n03163222}

\begin{figure*}[!h]\centering

    \begin{subfigure}[b]{\cmwidth}
    \includegraphics[width=\textwidth]{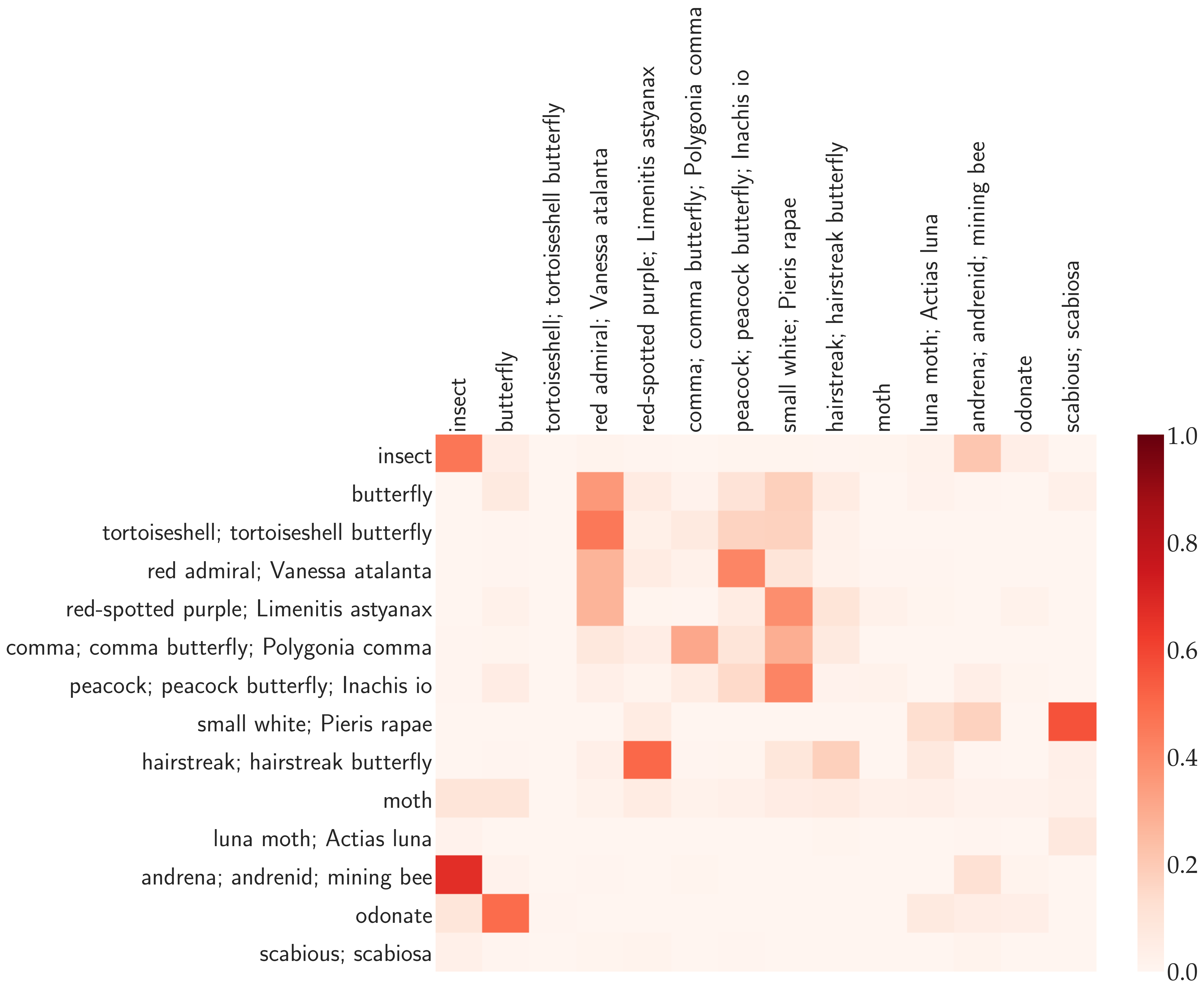}
    \caption{word2vec - class names}
    \end{subfigure}
    \begin{subfigure}[b]{\cmwidth}
    \includegraphics[width=\textwidth]{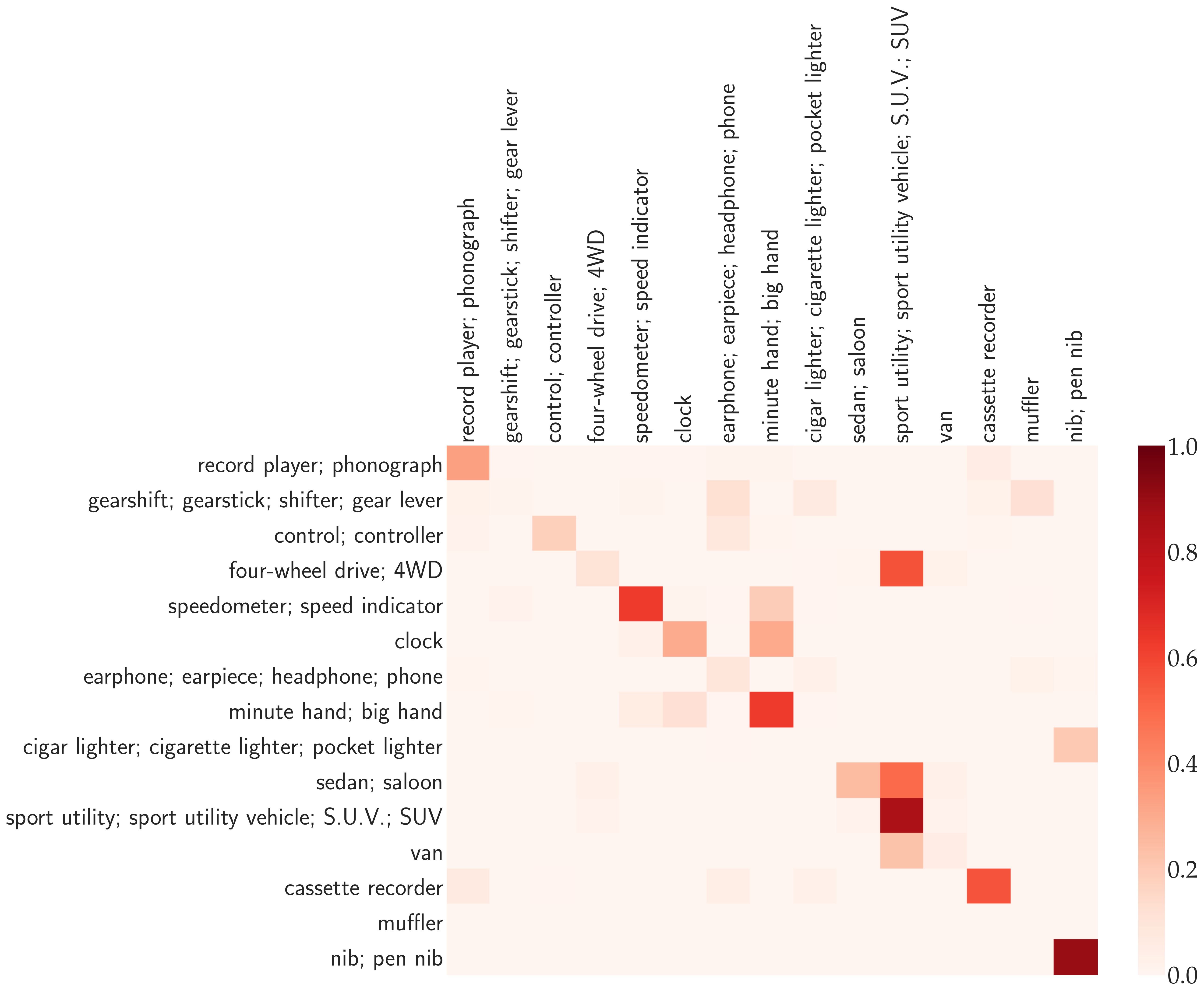}
    \caption{word2vec - class names}
    \end{subfigure}
    
    \begin{subfigure}[b]{\cmwidth}
    \includegraphics[width=\textwidth]{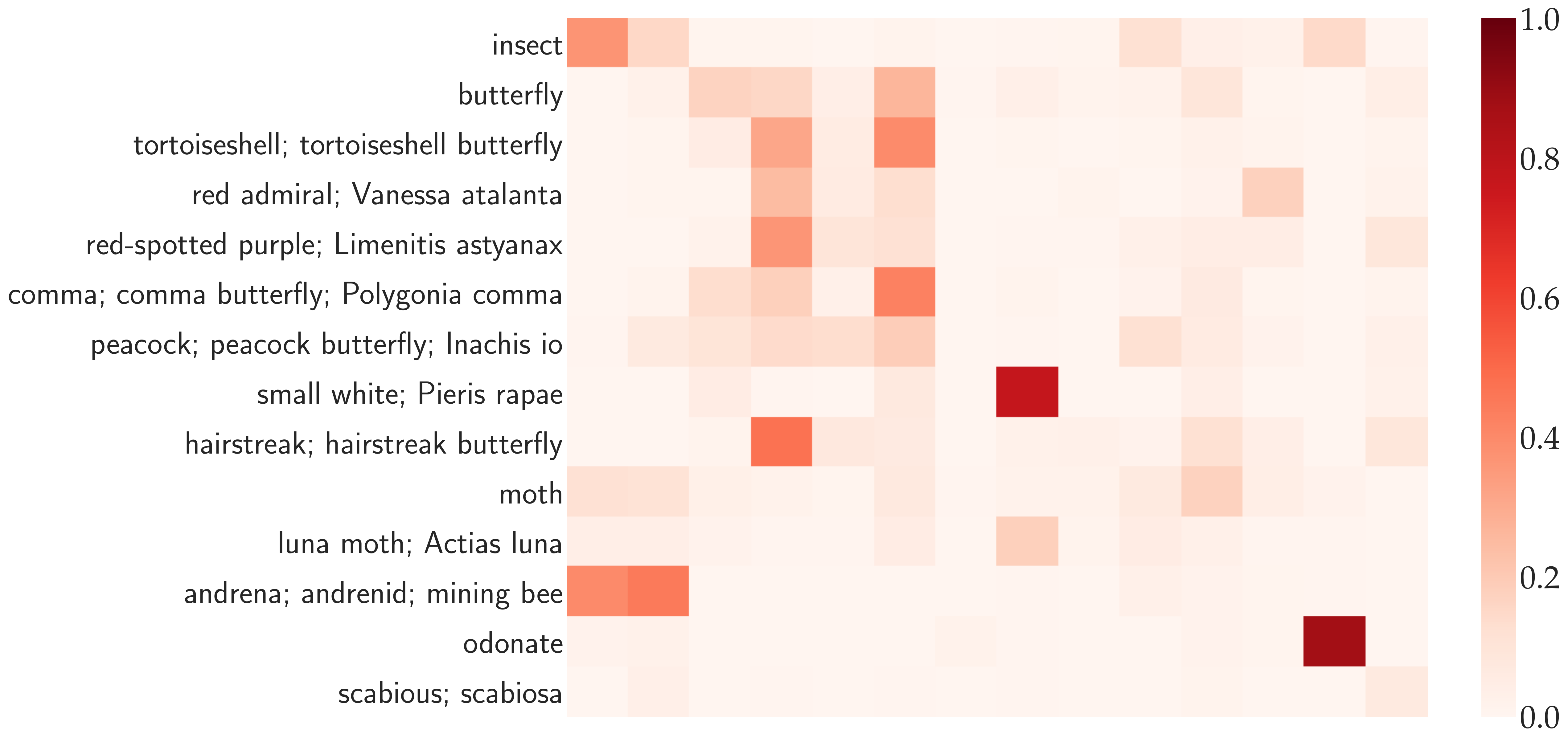}
    \caption{GloVe - Wiki articles}
    \end{subfigure}
    \begin{subfigure}[b]{\cmwidth}
    \includegraphics[width=\textwidth]{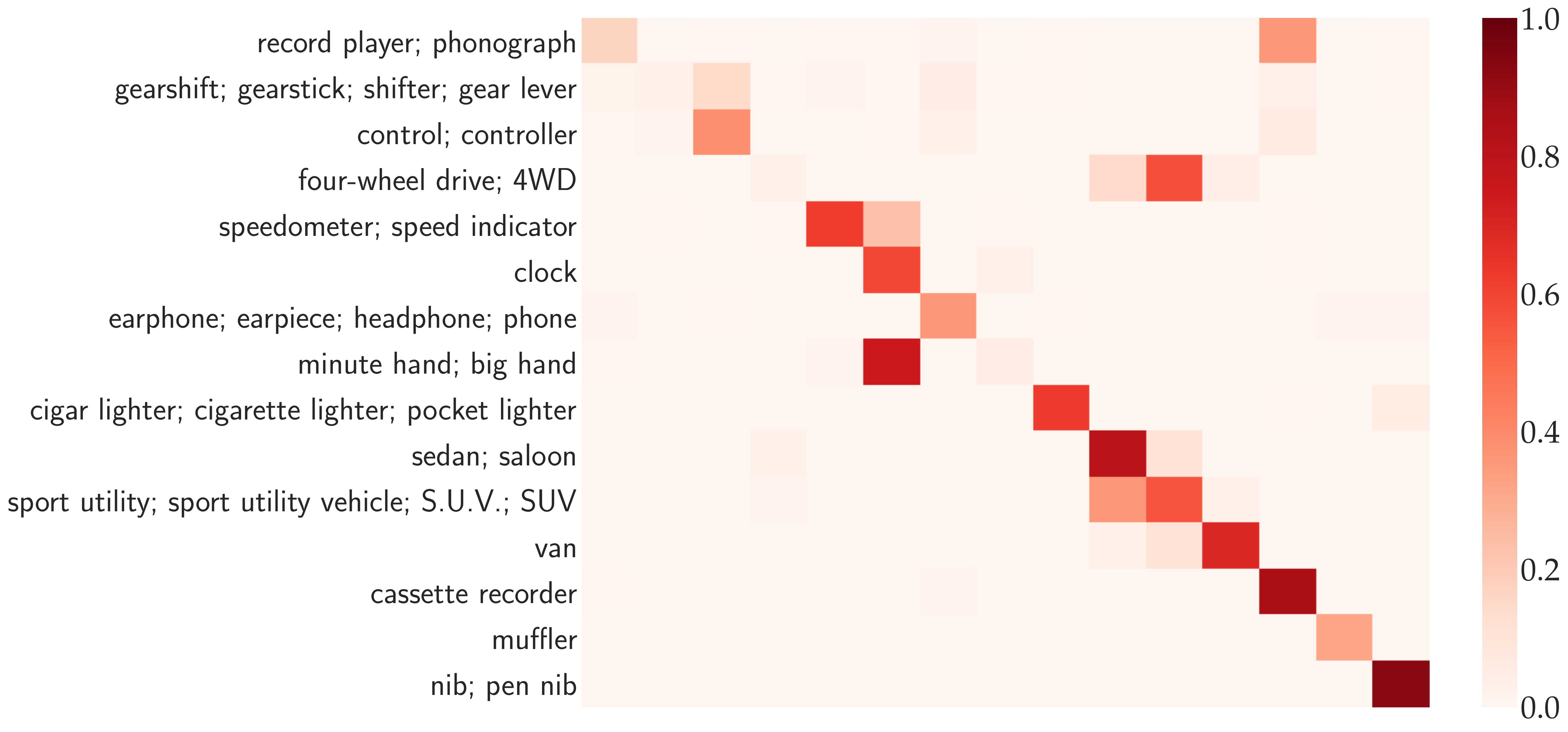}
    \caption{GloVe - Wiki articles}
    \end{subfigure}
    
    \begin{subfigure}[b]{\cmwidth}
    \includegraphics[width=\textwidth]{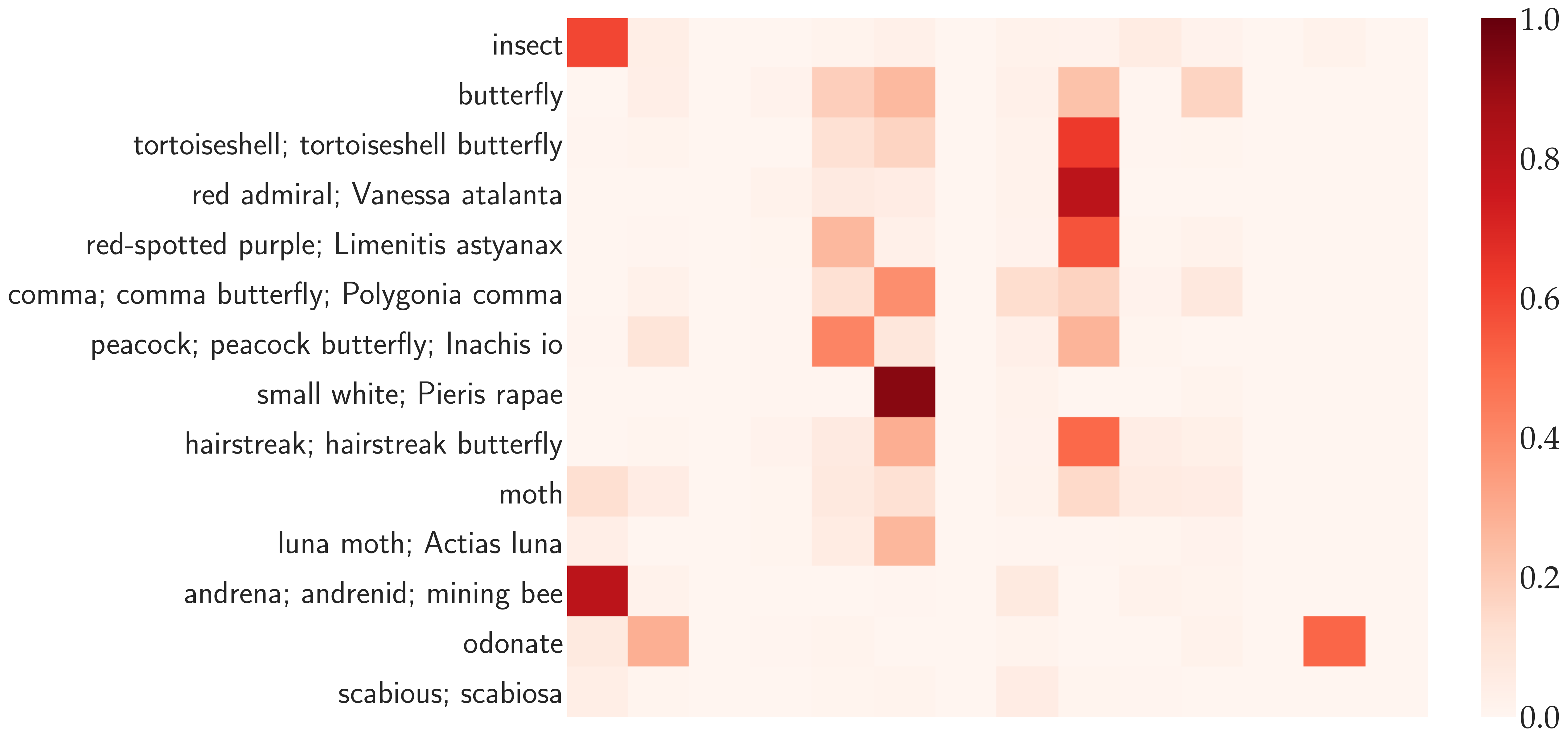}
    \caption{ALBERT-xxlarge - Wiki articles}
    \end{subfigure}
    \begin{subfigure}[b]{\cmwidth}
    \includegraphics[width=\textwidth]{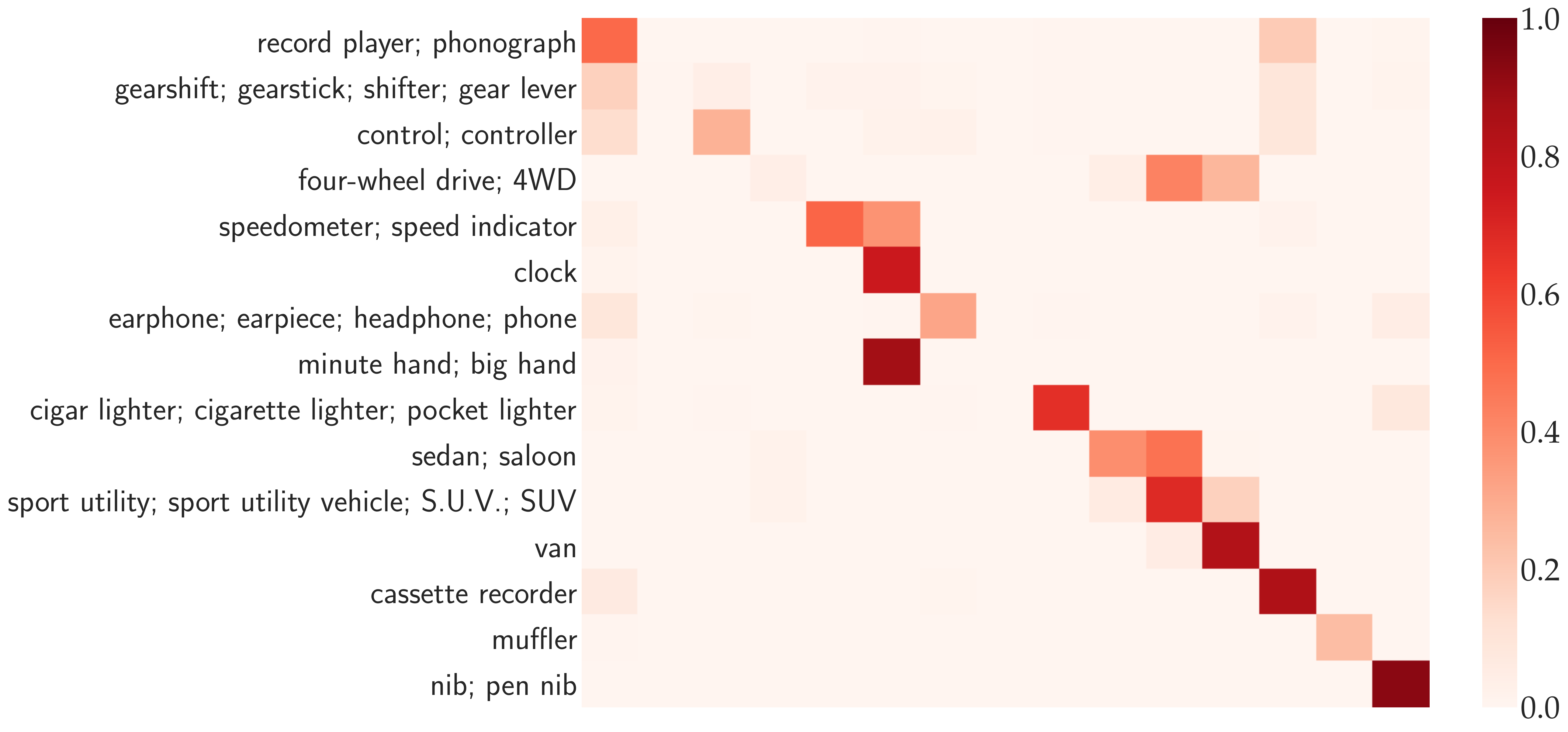}
    \caption{ALBERT-xxlarge - Wiki articles}
    \end{subfigure}

\caption{
\textbf{Subsets of confusion matrices on mp500 test set.}
All the models are CADA-VAE~\cite{schonfeld2019generalized} trained on different types of auxiliary data.
The rows correspond to true labels, columns to predicted classes - the matrices are row-normalized.
The subset of classes was selected by choosing one class and including others that were the most frequently mutually confused by any of the models.
The general performance differs between different subsets of classes - the pattern is less diagonal on the left than on the right column.
}
\label{fig:qualitative_confusion_matrices_appendix}
\end{figure*}

\section{Additional confusion matrices}
\label{app:confusion_matrices}
In Figure~\ref{fig:qualitative_confusion_matrices_appendix}, we show confusion matrices for more subsets of classes. On one of the subsets (left), the predictions are generally more noisy. However, the mispredictions generally fall within similar classes (different species of insects on the left side).
Additionally, we include the full confusion matrices in the supplementary material as standalone files.

\section{Additional across-category generalization results}
\begin{figure*}[!h]
\begin{subfigure}[b]{\textwidth}
     \def\numsize{\scriptsize}
\resizebox{\textwidth}{!}{
\begin{tikzpicture}[>=latex]
  
  \begin{axis}[
      width=24cm,
      height=8cm,
    ybar,
    ymin=0, ymax=79,
    xmin=.5, xmax = 9.6,
    enlarge x limits=0.02,    
    bar width=0.26cm,
    ylabel style={at={(current axis.above origin)},anchor=north west, yshift=-40pt, xshift=8pt, rotate=-90},
    xlabel style={at={(current axis.east)},anchor=east, yshift=-100pt, xshift=90pt,align=left},
    legend style={anchor=north west,legend columns=1, column sep=5pt,font=\footnotesize},
    legend cell align={left},
    ybar legend,        
    ylabel={\textbf{Top-5 accuracy} (\%)},
    xlabel = {\textcolor{gray}{\textbf{encoding \&}}\\ \textcolor{gray}{\textbf{auxiliary data}} \\[5pt] \textbf{Test class category} 
    \\ {\small{(number of classes)}}
    }, 
    axis x line=bottom,
    axis y line=left,    
    xtick={1,2,3, 4, 5, 6, 7, 8, 9},
    xticklabels = {\textbf{ALBERT}\\[1pt] \textbf{Wiki article}, \textbf{GloVe} \\[1pt] \textbf{Wiki article}, \textbf{word2vec}*\\[1pt] \textbf{Class Name}, \textbf{ALBERT}\\[1pt] \textbf{Wiki article}, \textbf{GloVe} \\[1pt] \textbf{Wiki article}, \textbf{word2vec}*\\[1pt] \textbf{Class Name}, \textbf{ALBERT}\\[1pt] \textbf{Wiki article}, \textbf{GloVe} \\[1pt] \textbf{Wiki article}, \textbf{word2vec}*\\[1pt] \textbf{Class Name}},
    x tick label style={font=\footnotesize, align=center},        
    axis line style={-Latex[round],thick},
    area style,
    every node near coord/.append style={rotate=90, anchor=north west,font=\scriptsize, yshift=5.5pt, xshift=0pt},
    visualization depends on={y\as\YY},
    nodes near coords={\pgfmathtruncatemacro{\YY}{ifthenelse(\YY==0,0,1)}\ifnum\YY=0\else\pgfmathprintnumber\pgfplotspointmeta\fi}        
    ]

    \begin{scope}[on background layer]
      \fill[black!20,opacity=1] ({rel axis cs:.01,-.14}) rectangle ({rel axis cs:0.325,0});
      \fill[black!20,opacity=1] ({rel axis cs:.335,-.14}) rectangle ({rel axis cs:0.655,0});
      \fill[black!20,opacity=1] ({rel axis cs:.67,-.14}) rectangle ({rel axis cs:0.985,0});      
      \fill[black!40,opacity=1] ({rel axis cs:.01,-.26}) rectangle
      node[black]{\large \textbf{Animals} {\small (82, 84*)}} ({rel axis cs:0.325,-.14});
      \fill[black!40,opacity=1] ({rel axis cs:.335,-.26}) rectangle node[black]{\textbf{\large Plants} {\small (39, 40*)}} ({rel axis cs:0.655,-.14});
      \fill[black!40,opacity=1] ({rel axis cs:.67,-.26}) rectangle node[black]{\textbf{\large Remaining} {\small (368, 376*)}} ({rel axis cs:0.985,-.14});            
    \end{scope}
    
  \addlegendimage{empty legend}

\addplot coordinates {
(1, 50.67) (4, 15.48) (7, 44.93)        % ALBERT
(2, 51.13) (5, 19.83) (8, 47.72)        % GloVe
(3, 33.49) (6, 12.76) (9, 27.72)        % W2V
};
\addplot coordinates {
(1,  3.65) (4, 16.69) (7, 43.48)        % ALBERT
(2,  5.46) (5, 17.76) (8, 45.82)        % GloVe
(3, 10.57) (6, 12.73) (9, 24.89)        % W2V
};
\addplot coordinates {
(1, 47.44) (4, 13.66) (7, 45.13)        % ALBERT
(2, 51.09) (5, 18.36) (8, 47.61)        % GloVe
(3, 30.60) (6, 12.97) (9, 26.18)        % W2V
};
\addplot coordinates {
(1, 47.52) (4, 18.38) (7, 30.24)        % ALBERT
(2, 47.89) (5, 18.30) (8, 34.62)        % GloVe
(3, 30.75) (6, 16.20) (9, 19.65)        % W2V
};
\addplot coordinates {
(1, 47.92) (4, 14.88) (7, 44.97)        % ALBERT
(2, 48.33) (5, 20.14) (8, 47.58)        % GloVe
(3, 33.41) (6, 13.31) (9, 26.79)        % W2V
};

\addlegendentry{\hspace{-25pt}\textbf{\footnotesize Training classes excluded:}}
\addlegendentry{None (0)}
\addlegendentry{All animal (398)}
\addlegendentry{All plant (31)}
\addlegendentry{Random non-animal (398)}
\addlegendentry{Random non-plant (31)}
\end{axis}
\end{tikzpicture} 
}
    \caption{\footnotesize{Simple ZSL, Top-5 accuracy}}
\end{subfigure}
\begin{subfigure}[b]{\textwidth}
     \def\numsize{\scriptsize}
\resizebox{\textwidth}{!}{
\begin{tikzpicture}[>=latex]
  
  \begin{axis}[
      width=24cm,
      height=8cm,
    ybar,
    ymin=0, ymax=37,
    xmin=.5, xmax = 9.6,
    enlarge x limits=0.02,    
    bar width=0.26cm,
    ylabel style={at={(current axis.above origin)},anchor=north west, yshift=-40pt, xshift=8pt, rotate=-90},
    xlabel style={at={(current axis.east)},anchor=east, yshift=-100pt, xshift=90pt,align=left},
    legend style={anchor=north west,legend columns=1, column sep=5pt,font=\footnotesize},
    legend cell align={left},
    ybar legend,        
    ylabel={\textbf{Top-1 accuracy} (\%)},
    xlabel = {\textcolor{gray}{\textbf{encoding \&}}\\ \textcolor{gray}{\textbf{auxiliary data}} \\[5pt] \textbf{Test class category} 
    \\ {\small{(number of classes)}}
    }, 
    axis x line=bottom,
    axis y line=left,    
    xtick={1,2,3, 4, 5, 6, 7, 8, 9},
    xticklabels = {\textbf{ALBERT}\\[1pt] \textbf{Wiki article}, \textbf{GloVe} \\[1pt] \textbf{Wiki article}, \textbf{word2vec}*\\[1pt] \textbf{Class Name}, \textbf{ALBERT}\\[1pt] \textbf{Wiki article}, \textbf{GloVe} \\[1pt] \textbf{Wiki article}, \textbf{word2vec}*\\[1pt] \textbf{Class Name}, \textbf{ALBERT}\\[1pt] \textbf{Wiki article}, \textbf{GloVe} \\[1pt] \textbf{Wiki article}, \textbf{word2vec}*\\[1pt] \textbf{Class Name}},
    x tick label style={font=\footnotesize, align=center},        
    axis line style={-Latex[round],thick},
    area style,
    every node near coord/.append style={rotate=90, anchor=north west,font=\scriptsize, yshift=5.5pt, xshift=0pt},
    visualization depends on={y\as\YY},
    nodes near coords={\pgfmathtruncatemacro{\YY}{ifthenelse(\YY==0,0,1)}\ifnum\YY=0\else\pgfmathprintnumber\pgfplotspointmeta\fi}        
    ]

    \begin{scope}[on background layer]
      \fill[black!20,opacity=1] ({rel axis cs:.01,-.14}) rectangle ({rel axis cs:0.325,0});
      \fill[black!20,opacity=1] ({rel axis cs:.335,-.14}) rectangle ({rel axis cs:0.655,0});
      \fill[black!20,opacity=1] ({rel axis cs:.67,-.14}) rectangle ({rel axis cs:0.985,0});      
      \fill[black!40,opacity=1] ({rel axis cs:.01,-.26}) rectangle
      node[black]{\large \textbf{Animals} {\small (82, 84*)}} ({rel axis cs:0.325,-.14});
      \fill[black!40,opacity=1] ({rel axis cs:.335,-.26}) rectangle node[black]{\textbf{\large Plants} {\small (39, 40*)}} ({rel axis cs:0.655,-.14});
      \fill[black!40,opacity=1] ({rel axis cs:.67,-.26}) rectangle node[black]{\textbf{\large Remaining} {\small (368, 376*)}} ({rel axis cs:0.985,-.14});            
    \end{scope}
    
  \addlegendimage{empty legend}

\addplot coordinates {
(1, 23.84) (4,  3.43) (7, 19.78)        % ALBERT
(2, 30.41) (5,  5.32) (8, 22.90)        % GloVe
(3, 24.30) (6,  4.47) (9, 15.24)        % W2V
};
\addplot coordinates {
(1,  1.94) (4,  3.69) (7, 20.67)        % ALBERT
(2,  3.52) (5,  4.47) (8, 23.36)        % GloVe
(3,  1.88) (6,  5.99) (9, 15.39)        % W2V
};
\addplot coordinates {
(1, 24.84) (4,  1.07) (7, 19.88)        % ALBERT
(2, 29.56) (5,  3.93) (8, 22.64)        % GloVe
(3, 23.69) (6,  3.40) (9, 15.33)        % W2V
};
\addplot coordinates {
(1, 22.54) (4,  2.69) (7, 11.52)        % ALBERT
(2, 28.74) (5,  5.59) (8, 15.32)        % GloVe
(3, 23.64) (6,  4.07) (9, 10.64)        % W2V
};
\addplot coordinates {
(1, 24.53) (4,  3.05) (7, 19.88)        % ALBERT
(2, 28.29) (5,  6.01) (8, 22.61)        % GloVe
(3, 24.34) (6,  4.12) (9, 14.88)        % W2V
};

\addlegendentry{\hspace{-25pt}\textbf{\footnotesize Training classes excluded:}}
\addlegendentry{None (0)}
\addlegendentry{All animal (398)}
\addlegendentry{All plant (31)}
\addlegendentry{Random non-animal (398)}
\addlegendentry{Random non-plant (31)}
\end{axis}
\end{tikzpicture} 
}
    \caption{\footnotesize{CADA-VAE, Top-1 accuracy}}
\end{subfigure}
\begin{subfigure}[b]{\textwidth}
     \def\numsize{\scriptsize}
\resizebox{\textwidth}{!}{
\begin{tikzpicture}[>=latex]
  
  \begin{axis}[
      width=24cm,
      height=8cm,
    ybar,
    ymin=0, ymax=37,
    xmin=.5, xmax = 9.6,
    enlarge x limits=0.02,    
    bar width=0.26cm,
    ylabel style={at={(current axis.above origin)},anchor=north west, yshift=-40pt, xshift=8pt, rotate=-90},
    xlabel style={at={(current axis.east)},anchor=east, yshift=-100pt, xshift=90pt,align=left},
    legend style={anchor=north west,legend columns=1, column sep=5pt,font=\footnotesize},
    legend cell align={left},
    ybar legend,        
    ylabel={\textbf{Top-1 accuracy} (\%)},
    xlabel = {\textcolor{gray}{\textbf{encoding \&}}\\ \textcolor{gray}{\textbf{auxiliary data}} \\[5pt] \textbf{Test class category} 
    \\ {\small{(number of classes)}}
    }, 
    axis x line=bottom,
    axis y line=left,    
    xtick={1,2,3, 4, 5, 6, 7, 8, 9},
    xticklabels = {\textbf{ALBERT}\\[1pt] \textbf{Wiki article}, \textbf{GloVe} \\[1pt] \textbf{Wiki article}, \textbf{word2vec}*\\[1pt] \textbf{Class Name}, \textbf{ALBERT}\\[1pt] \textbf{Wiki article}, \textbf{GloVe} \\[1pt] \textbf{Wiki article}, \textbf{word2vec}*\\[1pt] \textbf{Class Name}, \textbf{ALBERT}\\[1pt] \textbf{Wiki article}, \textbf{GloVe} \\[1pt] \textbf{Wiki article}, \textbf{word2vec}*\\[1pt] \textbf{Class Name}},
    x tick label style={font=\footnotesize, align=center},        
    axis line style={-Latex[round],thick},
    area style,
    every node near coord/.append style={rotate=90, anchor=north west,font=\scriptsize, yshift=5.5pt, xshift=0pt},
    visualization depends on={y\as\YY},
    nodes near coords={\pgfmathtruncatemacro{\YY}{ifthenelse(\YY==0,0,1)}\ifnum\YY=0\else\pgfmathprintnumber\pgfplotspointmeta\fi}        
    ]

    \begin{scope}[on background layer]
      \fill[black!20,opacity=1] ({rel axis cs:.01,-.14}) rectangle ({rel axis cs:0.325,0});
      \fill[black!20,opacity=1] ({rel axis cs:.335,-.14}) rectangle ({rel axis cs:0.655,0});
      \fill[black!20,opacity=1] ({rel axis cs:.67,-.14}) rectangle ({rel axis cs:0.985,0});      
      \fill[black!40,opacity=1] ({rel axis cs:.01,-.26}) rectangle
      node[black]{\large \textbf{Animals} {\small (82, 84*)}} ({rel axis cs:0.325,-.14});
      \fill[black!40,opacity=1] ({rel axis cs:.335,-.26}) rectangle node[black]{\textbf{\large Plants} {\small (39, 40*)}} ({rel axis cs:0.655,-.14});
      \fill[black!40,opacity=1] ({rel axis cs:.67,-.26}) rectangle node[black]{\textbf{\large Remaining} {\small (368, 376*)}} ({rel axis cs:0.985,-.14});            
    \end{scope}
    
  \addlegendimage{empty legend}

\addplot coordinates {
(1, 17.74) (4,  4.51) (7, 18.25)        % ALBERT
(2, 16.07) (5,  4.74) (8, 15.16)        % GloVe
(3, 13.17) (6,  3.09) (9,  8.91)        % W2V
};
\addplot coordinates {
(1,  5.23) (4,  4.50) (7, 17.18)        % ALBERT
(2,  1.52) (5,  4.17) (8, 16.08)        % GloVe
(3,  2.39) (6,  3.55) (9,  8.02)        % W2V
};
\addplot coordinates {
(1, 16.42) (4,  2.93) (7, 18.04)        % ALBERT
(2, 15.68) (5,  3.75) (8, 18.33)        % GloVe
(3, 12.13) (6,  2.80) (9,  8.29)        % W2V
};
\addplot coordinates {
(1, 15.60) (4,  4.14) (7,  9.23)        % ALBERT
(2, 14.75) (5,  4.18) (8, 11.35)        % GloVe
(3, 11.27) (6,  4.30) (9,  5.82)        % W2V
};
\addplot coordinates {
(1, 17.63) (4,  4.53) (7, 18.03)        % ALBERT
(2, 14.20) (5,  4.31) (8, 18.17)        % GloVe
(3, 12.29) (6,  3.00) (9,  8.71)        % W2V
};

\addlegendentry{\hspace{-25pt}\textbf{\footnotesize Training classes excluded:}}
\addlegendentry{None (0)}
\addlegendentry{All animal (398)}
\addlegendentry{All plant (31)}
\addlegendentry{Random non-animal (398)}
\addlegendentry{Random non-plant (31)}
\end{axis}
\end{tikzpicture} 
}
    \caption{\footnotesize{Simple ZSL, Top-1 accuracy}}
\end{subfigure}
\caption[]{\textbf{Within and across category generalization of ZSL.}
   We show the effect of excluding a category of classes from
   the training set on ZSL performance for all mp500 test set categories. Specifically we compare excluding all \textit{animal} classes vs. excluding the
  same number of random \textit{non-animal} classes. We repeat the same process
  for the \textit{plant} class.
  Different models trained on different auxiliary data are compared.
  The performance numbers are from a single run each. The performance on the unseen \textit{animal} classes drops
  dramatically when the animal classes are removed from the training set, the red bars in the first grouping of plots. The same trend, though not as pronounced, can be seen for the \textit{remaining} classes, the gray bars in the last grouping of plots.
  The number of test classes is slightly higher for models using class name data (marked with *) since Wiki articles are missing for some classes.
  }
  \label{fig:exclude_groups_top1}
\end{figure*}
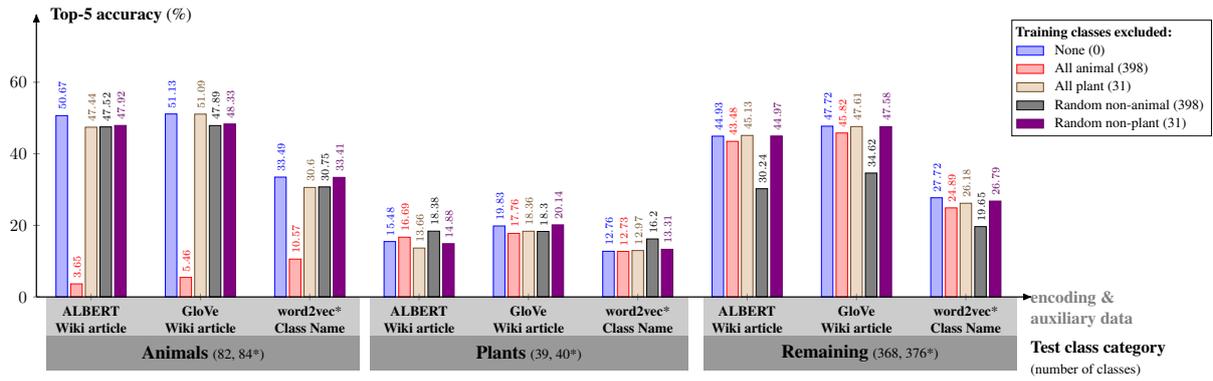
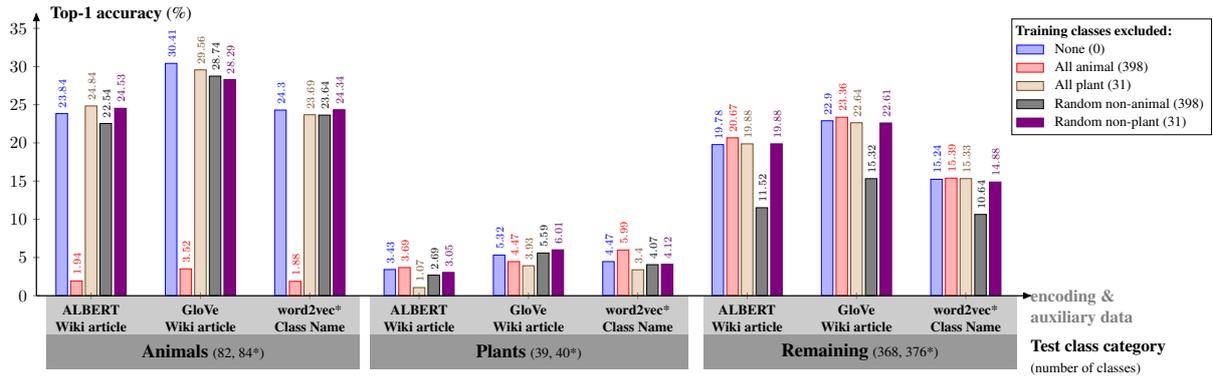
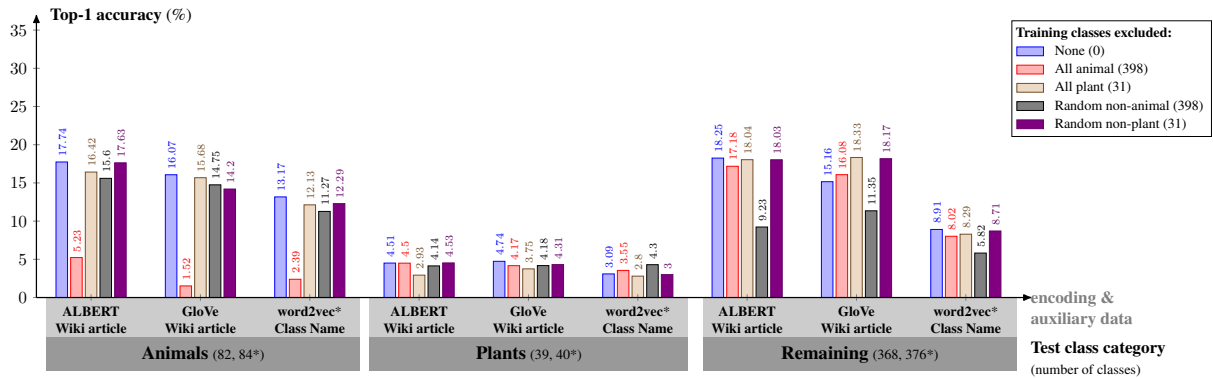

For completeness, in Figure~\ref{fig:exclude_groups_top1}, we also show the additional results of models trained on datasets that excluded different groups of classes.
The main trend is, however, similar to the observed in Figure~\ref{fig:exclude_groups}. We observe that models perform very poorly on animal classes if the training set excluded animals. Additionally, the performance on plant classes is generally lower - likely due to the smaller set of classes in the training set.

\begin{table*}[t]\centering
\caption{%
    \textbf{Validation set comparison of different textual inputs and encodings for zero-shot learning on ImageNet}.
    The reported metrics are mean per class top-1 and top-5 accuracies
    of the validation  classes. 750 \textit{train} classes were used as seen and 250 \textit{val} classes as unseen.
    All other hyparameters of Wikipedia models are the same and were chose based on the performance of the model from the first row. The model using class names had hyperparameters tuned independently.
    [CLS] represents the special output in ALBERT, typically used for classification tasks.
}
\ra{1.3}
{%
\begin{tabular}{@{}*{4}{l}*{2}{r}@{}}
  \toprule
  & & & & \multicolumn{2}{c}{\textbf{Result}}\\
  \cmidrule{5-6}
  \textbf{Auxiliary data} & \textbf{Features} & \textbf{Page rep.}& \textbf{Page aggreg.} & top-1 (\%)& top-5 (\%)\\
  \midrule
  Wiki articles & ALBERT ({\footnotesize xxlarge}) & mean tokens & mean & {\bftab 31.29} & {\bftab 64.92}\\
  Wiki articles & ALBERT ({\footnotesize xxlarge}) & mean tokens & sum & 29.77 & 62.44\\
  Wiki articles & ALBERT ({\footnotesize xxlarge}) & [CLS] output & mean & 3.24 & 12.45\\
  Class names & word2vec & N/A & N/A & 27.87 & 61.91\\
  \bottomrule
\end{tabular}
}
\label{tbl:val_results}
\end{table*}

\section{Different ALBERT encodings}
Table~\ref{tbl:val_results} compares different ways to encode the textual descriptions with ALBERT~\cite{lan2019albert}, such as how to extract features from individual pages and aggregate the page features for classes with more than one corresponding Wikipedia article.
ALBERT~\cite{lan2019albert} has a special output used for classification (``[CLS]''), which we additionally evaluate for extracting features. However, averaging features over each tokens seem to work much better. Additionally, averaging features among multiple pages works best.
All other our experiments use averaging both the page representations and for aggregating page features (as in row 1, Table~\ref{tbl:val_results}).

\label{app:different_albert_encodings}

\section{Model details}
\label{app:model_details}
\subsection{Simple ZSL}

We individually tune model hyperparameters for three different types of auxiliary data: class names with word2vec features, Wikipedia articles with either ALBERT-base or ALBERT-xxlarge features.
We use Adam optimizer and a random search to generate 40 different hyperparameter settings and choose the best setting for each model variant.
The ranges of values used for different hyperparameters were as following:
batch size $\in \{ 32, 128, 256, 512,  1024 \}$,
target projection dimension defined by the shapes of $W_t$ and $W_x$ $\in \{ 32, 128, 256, 512, 1024 \}$,
margin $m$ $\in (0.1, 100.)$ or with equal probability $m = 1$,
$\beta_1$ (for Adam) $\in \{0.5, 0.9\}$
learning rate $\in (0.00003, 0.01)$.
The values are sampled uniformly in the $\log$ space of the ranges.

The best configurations of hyperparameter values for each model variant are attached in the supplementary material together with the model weights.

\subsection{CADA-VAE}
\label{app:model_details_cada}

\paragraph{Implementation details}
We use our own implementation of CADA-VAE~[35] and verify that it performs similarly to the original. However, we there are two modification that we do.
The official implementation of CADA-VAE\footnote{\url{https://github.com/edgarschnfld/CADA-VAE-PyTorch}} appears to incorrectly implement reparameterization trick of Variation Auto-Encoders~[11].
It misses the factor of $0.5$ when transforming $\log (\sigma^2)$ into $\sigma$.
Instead, we use $\sigma = \exp (0.5 \cdot \log (\sigma^2))$.
Also, unlike the original authors, instead of using a sum to aggregate loss function over samples in a batch, we use a mean.
However, this difference should not be important as it can be compensated for by different learning rate or scaling factors of the loss function.

\paragraph{Hyperparameters}
Like for Simple ZSL, the best values of hyperparameters are chosen individually for each of the three different auxiliary data sources.
However, for CADA-VAE we use a combination of manual and random search over hyperparameter values.

The random search consisted of 42 runs on each of the model variant using the following ranges:
batch size $\in \{ 32, 128, 256, 512, 1024 \}$,
VAE latent space dimension $\in \{32, 128, 256, 512, 1024 \}$,
VAE image features encoder $\in \{ [1560, 1560], [2048, 1024], [1560], [1024, 512] \}$,
VAE image features decoder $\in \{ [1660], [1024, 2048], [1560], [512] \}$,
VAE auxiliary features encoder $\in \{ [1560, 1560], [2048, 1024], [1560], [1024, 512] \}$,
VAE auxiliary features decoder $\in \{ [1660], [1024, 2048], [1560], [2048], [512] \}$,
$\beta$ factor $\in (0.1, 30)$ or with probability $0.3$ fixed $\beta = 1$,
cross reconstruction loss factor $\in (0.25, 50.)$,
distributional alignment loss factor $\in (0.25, 100.)$.
Same as for Simple ZSL, the values were sampled uniformly in the $\log$ space of the ranges.
The loss function factors were used for linearly increasing loss function coefficients. The schedules for the coefficients were the same as in the original work, that is:
$\beta$ (for scaling VAE KL-divergence loss) was increased from epoch 0 to 93 (or fixed $\beta = 1$, cross reconstruction loss from epoch 21 to 75, and distribution alignment from epoch 6 to 22.
The rest of the hyperparameters were constant for the random search. We used AMSGrad optimizer, learning rate of VAEs of $0.00015$.
For the linear classifier we use: learning rate of $0.001$, Adam optimizer, batch size of $32$ and $200$ sampled latent space vectors from each class.
Additionally, we tried 58 evaluation runs on the variant using word2vec features, and only a subset of those settings on the models using ALBERT-base (21 runs), or ALBERT-xxlarge (19 runs) features.

The best configurations of hyperparameter values for each model variant are attached in the supplementary material together with the model weights.

\end{document}